\documentclass[10pt,twocolumn,letterpaper]{article}

\usepackage{3dv}
\usepackage{times}
\usepackage{epsfig}
\usepackage{graphicx}
\usepackage{amsmath}
\usepackage{amssymb}

\usepackage{etoolbox}
\usepackage{silence}
\makeatletter
\robustify\@latex@warning@no@line
\makeatother

\usepackage{authblk}
\usepackage{caption}
\usepackage{times}
\usepackage{epsfig}
\usepackage{graphicx}
\usepackage{amsmath}
\usepackage{amssymb}
\usepackage{multirow}
\usepackage{booktabs}

\usepackage{pifont}
\usepackage[ruled,vlined]{algorithm2e}
\usepackage[font=small]{caption}

\usepackage[pagebackref=true,breaklinks=true,colorlinks,bookmarks=false]{hyperref}

\threedvfinalcopy 

\def\threedvPaperID{260} 

\ifthreedvfinal\fi

\newcommand{\ignore}[1]{}

\newcommand{\norm}[1]{\left\lVert#1\right\rVert}
\begin{document}

\title{Learning 3D Semantic Segmentation with only 2D Image Supervision}

\makeatletter
\renewcommand\Authfont{\fontsize{11.5}{14.4}\selectfont}
\renewcommand\AB@affilsepx{\qquad \protect\Affilfont}
\makeatother
\author[1,2]{Kyle Genova}
\author[1]{Xiaoqi Yin}
\author[1]{Abhijit Kundu}
\author[1]{Caroline Pantofaru}
\author[1]{Forrester Cole}
\author[1]{\\Avneesh Sud}
\author[1]{Brian Brewington}
\author[1]{Brian Shucker}
\author[1, 2]{Thomas Funkhouser}
\affil[1]{Google Research}
\affil[2]{Princeton University}
\renewcommand*{\Authsep}{ \ }
\renewcommand*{\Authands}{ \ }

\maketitle


\begin{abstract}
With the recent growth of urban mapping and autonomous driving efforts, there has been an explosion of raw 3D data collected from terrestrial platforms with lidar scanners and color cameras.  However, due to high labeling costs, ground-truth 3D semantic segmentation annotations are limited in both quantity and geographic diversity, while also being difficult to transfer across sensors. In contrast, large image collections with ground-truth semantic segmentations are readily available for diverse sets of scenes.   In this paper, we investigate how to use only those labeled 2D image collections to supervise training 3D semantic segmentation models.  Our approach is to train a 3D model from pseudo-labels derived from 2D semantic image segmentations using multiview fusion.  We address several novel issues with this approach, including how to select trusted pseudo-labels, how to sample 3D scenes with rare object categories, and how to decouple input features from 2D images from pseudo-labels during training.  The proposed network architecture, 2D3DNet, achieves significantly better performance (+6.2-11.4 mIoU) than baselines during experiments on a new urban dataset with lidar and images captured in 20 cities across 5 continents.
\end{abstract}

\section{Introduction}
\label{sec:intro}

\begin{figure}[t]
    \centering
    \includegraphics[width=0.9
    \columnwidth]{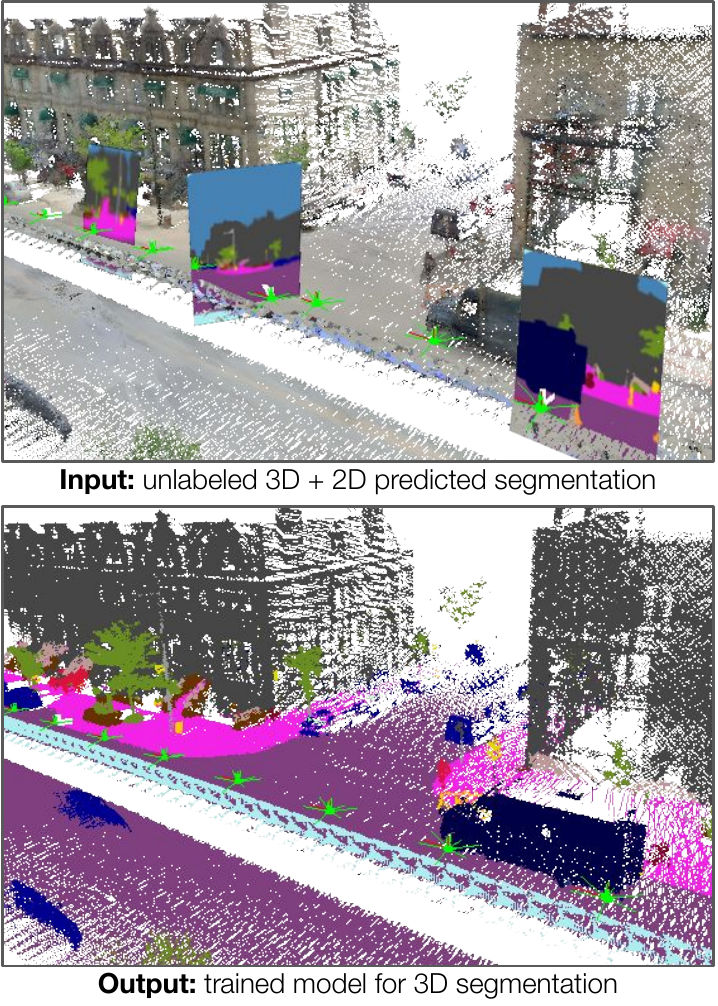}
    \caption{Semantic segmentation of 3D data without any 3D supervision.  We leverage semantic segmentations of posed 2D images (top) to produce pseudo-labels to train a model for semantic segmentation of 3D point clouds (bottom).}
    \label{fig:teaser}
    \vspace{-4mm}
\end{figure}

Semantic segmentation of 3D scenes is a fundamental task in computer vision with applications in semantic mapping, logistics planning, augmented reality, and several other domains.  In these applications, the input data is typically a sequence of laser scans and posed color images acquired from a terrestrial platform, and the task is to predict a category label for every 3D point (Figure \ref{fig:teaser}).  In this work, we aim for a method that generalizes to real-world data captured with any sensor anywhere in the world.

This task is challenging due in part to the difficulty of obtaining training datasets with dense 3D semantic annotations.   Several 3D semantic segmentation benchmarks have been released for outdoor scenes in the past few years \cite{behley2019semantickitti,caesar2019nuscenes,hackel2017isprs,pan2020semanticposs,PandaSet,roynard2018paris}.  However, they each contain data captured with a unique lidar sensor configuration in a unique part of the world.  As a result, training on one of these datasets and then inferring on another performs poorly, as models fit to the specific sampling patterns and unique content of their training data (see Appendix~\ref{sec:kitti_comparison}).  While one could try unsupervised domain adaptation, generalization to new lidar sensor configurations is notoriously difficult, and SOTA domain adaptation methods are not competitive for semantic segmentation of 3D point clouds \cite{yi2020complete}.

In this work, we aim for a 3D semantic segmentation method that works in-the-wild for data captured with any lidar sensor anywhere in the world, even where no 3D semantic annotations are available.   This focus is motivated by the needs of world mapping organizations, which continuously collect massive stores of unlabeled 3D lidar and image data from around the globe with a variety of sensors, but do not have ground-truth 3D annotations for it.

Our solution is to leverage labeled 2D image collections.   There are a number of 2D image datasets with per-pixel semantic labels \cite{cordts2016cityscapes,lin2014microsoft}, some of which are quite large and diverse, with images from around the world \cite{neuhold2017mapillary}.  Supervised 2D models trained on these datasets have been shown to generalize well to images from diverse settings~\cite{lambert2020mseg}.  Our approach is to use these trained 2D models to produce pseudo-labels for training 3D models.   In this way, we can train a 3D semantic segmentation model uniquely for each 3D dataset without requiring any 3D annotations.

\begin{figure*}
    \centering
    \vspace{-5em}
    \includegraphics[width=\textwidth]{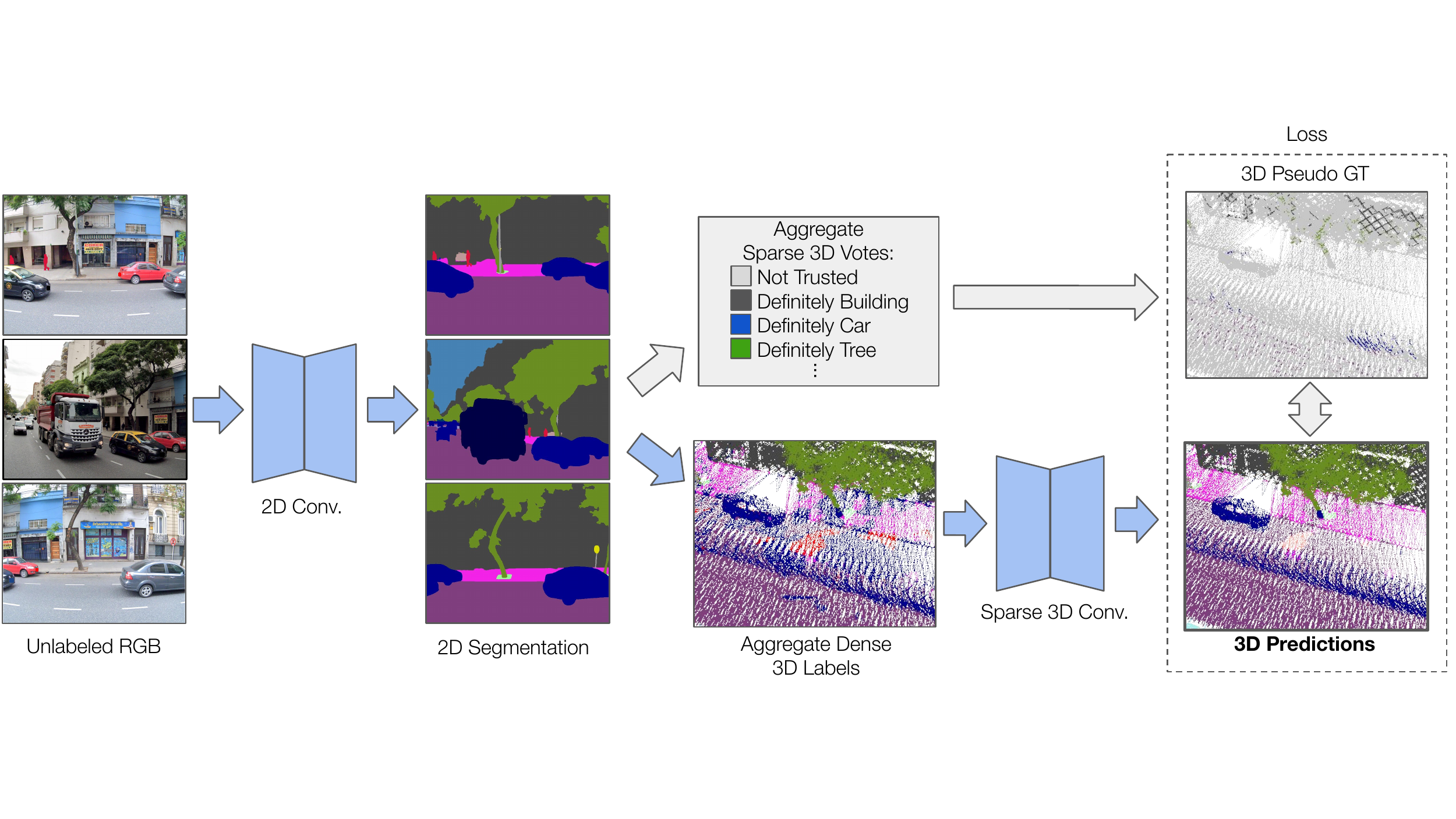}
    \vspace{-6em}
    \caption{Our approach is composed of three stages. First, we run a pretrained 2D segmentation network on the input RGB images. Second, we aggregate the images into both dense, noisy segmentations and sparse, clean segmentations.  Third, the dense segmentations are passed to a 3D sparse voxel convolution network, and a standard softmax cross-entropy loss is applied at the sparse locations.}
    \label{fig:method}
    \vspace{-4mm}
\end{figure*}

Our network architecture, which we call 2D3DNet, is shown in Figure \ref{fig:method}.   For each 3D scene in an unlabeled training set, we first run a pre-trained 2D model on every image to produce a semantic label for each pixel.   Then for each 3D point, we back-project it into every image to produce a set of weighted votes, which are tallied to produce features and also pseudo-labels.  Finally, we train a 3D model that takes in 3D points with features and outputs semantic labels using supervision from the pseudo-labels. At test time, we run the 2D model to produce features (optionally), and then the 3D model to predict a final 3D semantic segmentation. 

Although this basic method addresses the main issue (how to supervise 3D with 2D), it has several issues.   First, pseudo-labels can be very noisy, partly due to errors made by 2D models, and partly due to errors in 2D-3D correspondences. It is not enough to simply distill a 3D model from corresponding 2D pixel predictions (see Section~\ref{sec:comparisons}). Second, training a 3D network that takes in features from 2D images and predicts pseudolabels from the same images can create a correlation that impedes learning.  Of course, features from 2D images are critical to performing 3D semantic segmentation, since many classes are easier to discriminate in images than point clouds (see Table~\ref{tab:features} in the Appendix). Finally, because ground truth 3D annotations are not required, the potential training datasets are massive, raising interesting questions of how to sample training data.

We have addressed these issues by creating trusted sparse pseudo-labels, preventing feature coupling, and optimizing our dataset for rare and interesting examples. The result is a generalizable 3D model with supervision only from unpaired 2D images. We train the model on an unlabeled dataset of 250,000 scenes, and test it on a diverse set of scenes from 20 cities in five continents. This approach improves over baselines by 6.2-11.4 mIoU, and is competitive with state of the art methods on the nuScenes lidarseg benchmark while reducing its reliance on 3D annotations.

\ignore{
Our key contributions are as follows:

\begin{itemize}
\setlength{\topsep}{0pt}
\setlength{\parsep}{0pt}
\setlength{\partopsep}{0pt}
\setlength{\parskip}{0pt}
\setlength{\itemsep}{2pt}
\vspace*{-2mm}
  \item We train a 3D semantic segmentation network using only 2D supervision, overcoming challenges of applying multiview fusion to large-scale in-the-wild data.
    \item We show that creating sparse, high-quality supervision is preferable to training with dense supervision made from noisy 2D$\rightarrow$3D correspondences.
    \item We overcome the problems that arise when using features and supervision derived from the same 2D images. Learning in a way that generalizes to all points is a critical step to achieving good results.
    \item We demonstrate a diverse scene sampling approach that outperforms random sampling for the same number of training samples.
    \item We provide results of experiments showing our algorithms and network architecture trained with 2D and 3D supervision performs at SOTA on the nuScenes 3D Semantic Segmentation Task.
\end{itemize}
}


\section{Related Work}
\label{sec:related}

\vspace*{2mm}\noindent{\bf 3D semantic segmentation:} Many
researchers have worked on 3D semantic segmentation \cite{che2019object,guo2020deep,liu2019deep, cortinhal2020salsanext, milioto2019iros, zhang2020polarnet}.  For outdoor scenes, most early work focused on separately labeling terrain~\cite{Mallet08,Torre00,Wolf05},
roads~\cite{Clode04,Hatger03,Jaakkola08,Yu07,Zhao12a},
buildings~\cite{Bredif07,Chen08,Frueh05,Lafarge08,Ortner07}, trees~\cite{Wang08,Xu07}, and
stand-alone objects~\cite{golovinskiy2009shape}.  Recent work has focused on 3D network architectures to extract features from 3D point clouds \cite{wang2019towards, pham2019jsis3d,qi2017pointnet,qi2017pointnet++,shi2019pv,thomas2019kpconv,jsenet}, meshes \cite{hanocka2019meshcnn,huang2019texturenet}, voxels \cite{song2017semantic}, octrees \cite{riegler2017octnet}, and sparse grids \cite{choy20194d,choy2019fully,graham20183d,han2020occuseg}.  Yet, all of these methods require expensive labeled 3D data for training.

\vspace*{2mm}\noindent{\bf 3D semantic segmentation datasets.} Some datasets have been released with ground-truth 3D semantic segmentation labels in the past few years.  Examples include Paris-Lille-3D \cite{roynard2018paris} (2km in two cities in France),  SemanticKITTI \cite{behley2019semantickitti} (40km near Karlsruhe), Pandaset \cite{PandaSet} (a few sequences near San Francisco), and NuScenes Lidarseg \cite{caesar2019nuscenes} (1000 scenes from one neighborhood in Boston and three in Singapore).  These datasets lack diversity, with most covering one or two small regions.  Plus, they are all captured with different lidar configurations, which makes cross-dataset training and domain adaptation difficult \cite{achituve2021self,alonso2020domain,jaritz2021cross,jaritz2020xmuda,xu2020squeezesegv3}.   In contrast, labeled 2D datasets have great diversity.  For example, Mapillary Vistas has labeled images from 25,000 scenes across six continents \cite{neuhold2017mapillary}.   We leverage the size and diversity of image datasets to train generalizable 3D models.

\vspace*{2mm}\noindent{\bf Multi-view fusion.}  
Previous approaches have explored aggregating 2D image semantic features onto 3D points using weighted averaging \cite{hermans2014dense,kundu2020virtual,vineet2015incremental, armeni_iccv19}, CRFs \cite{mccormac2017semanticfusion}, label diffusion~\cite{mascaro2021diffuser}, and Bayesian fusion \cite{ma2017multi,vineet2015incremental,zhang2019large}.   However, these previous works do not extract any features in the 3D domain, and thus they under-perform at recognizing objects with distinctive 3D shapes.  Plus, they do not produce a trained 3D model that can be applied to novel scenes without posed images.

\vspace*{2mm}\noindent{\bf Hybrid 2D-3D convolutions.}
Some existing methods include 3D convolutions after multi-view fusion of 2D image features~\cite{jaritz2019multi, dai20183dmv, 6907298, vora2020pointpainting}.  These methods follow the same basic approach as ours.  However, they do not address the issues at the heart of our work.   They require full 3D supervision, they assume good correspondence between 2D and 3D, and they consider only datasets with small scenes \cite{6907298}, indoor settings \cite{dai20183dmv, jaritz2019multi}, or foreground objects \cite{vora2020pointpainting}.  These methods would not generalize or scale to outdoor scenes from anywhere in the world.

\vspace*{2mm}\noindent{\bf Cross-modal supervision.} This work sits within the area of cross-modal supervision, using information in one modality to train models in another modality through correspondence.  Another example of this approach is in the video domain, where temporal synchronization provides correspondences between audio and visual, e.g. \cite{owens2016ambient, zhao2018soundofpixels, korbar2018cooperative, gan2019self, aytar2016soundnet, ephrat2018looking, koepke2020sight, alwassel2020self} or text, speech and visual signals, e.g. \cite{nagrani2020speech2action}. There are also methods that transfer supervision or share information from image models to other domains-- depth and  RGB~\cite{gupta2016cross, meyerimproving}, image and video~\cite{girdhar2019distinit}, or image and 3D~\cite{tian2019contrastive, jing2020self, lawin2017deep}. However, these image-based transfer approaches rely on the assumption that accurate correspondence is known (e.g., paired images). Here, the correspondence between modalities is only a sparse, noisy estimate.

\section{Methods}
\label{sec:methods}

Our network architecture, 2D3DNet, is composed of three stages (Figure~\ref{fig:method}). It takes as input a 2D semantic segmentation model (``2D Conv.'') pretrained on a labeled image collection and a ``scene'' containing a short $\sim$10s sequence of unlabeled RGB images and lidar points captured contemporaneously, but asynchronously (Figure~\ref{fig:teaser}).   In the first stage, it uses the pretrained 2D model to create 2D segmentation predictions for all image pixels.  In the second stage, it uses multi-view fusion to make best-guess semantic labels for as many 3D points as possible via backprojection and voting from labels of the corresponding pixels.   In the third stage, those aggregated dense 3D labels are converted to input features at the 3D points and run through a sparse 3D convolutional network (see Appendix \ref{sec:appendix_architecture_details} for architecture) to predict the semantic label for every 3D point.

Given this inference model, the main challenge is how to train it.   Previous models of this type are trained with supervision from in-domain labeled 3D data.   However, most real-world scenarios will not have such data.   Instead, they have large repositories of unlabeled data.   To use unlabeled data to train the 3D model, we generate 3D pseudo-ground-truth training data from predicted 2D semantic segmentations (Section~\ref{sec:pseudolabels}), carefully decouple the 2D segmentation image input features from the 3D pseudo-labels to ensure the 3D model generalizes beyond the training data (Section~\ref{sec:decoupling}), and select unlabeled scenes judiciously so that the trained model performs well on rare semantic categories (Section~\ref{sec:sampling}). Remaining implementation details are provided in Appendix \ref{sec:implementation}.


\subsection{Generating 3D Pseudo-Ground-Truth}
\label{sec:pseudolabels}

Our first problem is to create pseudo-ground-truth labels for 3D points in the training set.  Given a scene with predicted 2D semantic labels for all pixels of all images, our goal is to produce labels for 3D points that can be used to train a 3D semantic segmentation model.

\begin{figure}
    \centering
    \includegraphics[width=0.48\textwidth]{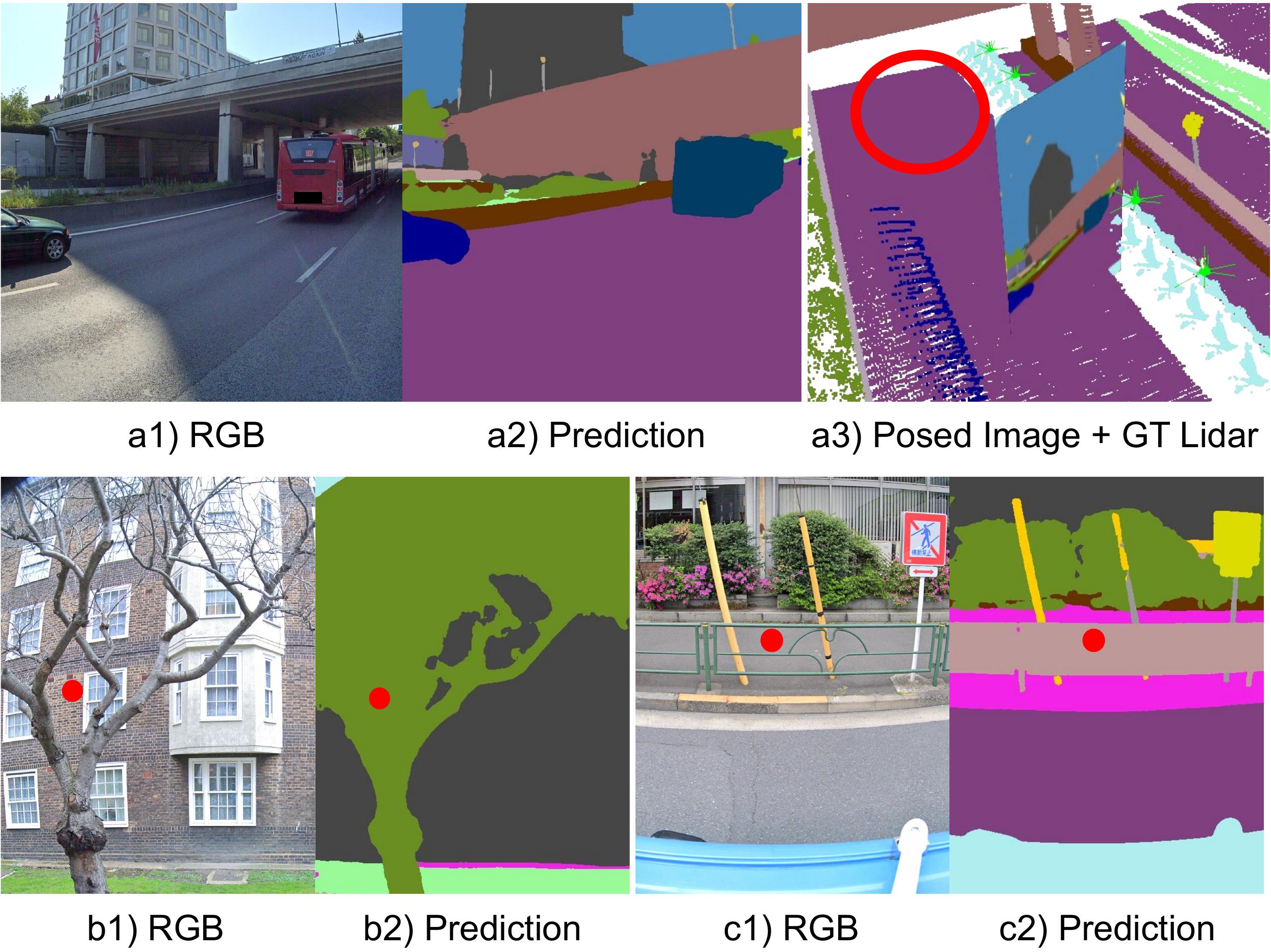}
    \vspace{-2em}
    \caption{Challenges of creating 3D supervision from 2D votes. (a) A car and a bus drive forward with the ego vehicle. Because of aliasing, the car has an extended lidar point cloud, while the bus is completely absent from the lidar (red circle). (b) The network frequently makes mistakes near occlusion boundaries, so points that are near the silhouette of an occluder are not trusted. (c) Labeling standards differ from 2D to 3D. In 2D, fences are typically filled completely, regardless of their density.}
    \label{fig:correspondenceproblem}
    \vspace{-3mm}
\end{figure}

Although it seems simple to backproject the predicted labels for all 2D pixels onto 3D points and vote, this na\"ive approach does not work well for several reasons (see results in Section~\ref{sec:comparisons}).
First, 3D lidar sweeps (10Hz) and 2D images (2Hz) are captured asynchronously with moving sensors, and thus there is no obvious one-to-one correspondence between 3D points and 2D pixels.   Second, real-world urban scenes contain dynamic objects that can move significantly between 2D image and 3D lidar observations (cars, people, etc.), and thus observations at time A by one sensor may be inconsistent with observations at time B by the other (Figure~\ref{fig:correspondenceproblem}a).  
Third, predicted semantic labels in 2D images are almost never pixel-perfect, with labels that bleed across object boundaries and lead to incorrect backprojections near occlusion boundaries (Figure~\ref{fig:correspondenceproblem}b).
Fourth, ground truth standards between 2D and 3D dataset annotation differ, for example fences are mostly see-through but usually segmented as solid in 2D for practical reasons (see Figure~\ref{fig:correspondenceproblem}c), leading to  inaccurate backprojection to 3D.

Our solution is to create pseudo-ground-truth for only the \emph{extremely sparse} set of 3D points for which 2D-3D correspondences can be established robustly.   The key observation is that unlabeled scene data is abundant, so we can throw out labels for 99\% or more of the 3D points and still have more than enough data to train a 3D model.   The goal then is to find the point-pixel correspondences we should trust. To implement this idea, we use a series of filters to exclude questionable point-pixel correspondences.

\vspace*{2mm}\noindent{\bf Temporal Filter}. First, we exclude points whose timestamps are more than $\Delta t_{max}$ from that of the closest image (by default $\Delta t_{max}=0.1$ seconds).   Since images are captured sparsely in time (2Hz in our data), this filter removes the majority of the lidar points. It also removes the most egregious types of correspondence errors caused by moving objects.   For example, in Figure \ref{fig:correspondenceproblem}a, points on the road behind the moving bus observed before or after the bus drove by will not be matched with the ``bus'' pixels in the image. 

\vspace*{2mm}\noindent{\bf Occlusion Filter}. Second, we apply an occlusion test that conservatively over-estimates the extent of foreground objects. To check for occlusion, we render a depth channel per image where the lidar points are drawn as discs (surfels~\cite{pfister2000surfels}) with normals and two radii expressing the size and aspect ratio. The normals and radii are estimated per surfel by computing the size and aspect ratio of the local neighborhood of points using PCA (see Appendix~\ref{sec:normal_estimation} for details). After rendering depth images, a point is considered occluded if its projected depth mismatches the depth image by a factor exceeding $\tau=1\%$. To conservatively over-estimate the extent of foreground objects, we purposely multiply (``dilate'') the base estimated radii of the lidar discs by a factor of $k=8$. A conservative test is needed for cases such as the red points in Figure~\ref{fig:correspondenceproblem}b,c, which would otherwise receive incorrect predicted class labels because of imperfect segmentation masks.

\vspace*{2mm}\noindent{\bf Weighted Voting}.
Third, we aggregate the 2D predicted labels that passed the filters to determine the pseudo-ground-truth 3D point labels. The scheme up-weights near-in-time, near-to-camera views while still including more spatially or temporally distant observations if they are all that is available. Concretely, given thresholds for the maximum distance $d_{max}$ and maximum timestamp difference $\Delta t_{max}$, the weight for an image-$i$, point-$j$ pair is:

\[
w_{ij} = \left(1 - \frac{\|p^{cam}_i - p^{pt}_{j}\|^2_2}{d^2_{max}}\right)^2\left(1 - \frac{|t^{cam}_i - t^{pt}_j|^2}{\Delta t^2_{max}}\right)^2
\]

In combination, these filters exclude almost all 3D points from the pseudo-ground-truth set, as designed.  By applying the filters to vast amounts of unlabeled data, we produce a very large training set with reliable pseudo-ground-truth.  Our filters are very conservative and thus produce few false positives that could affect the 3D network training process. 


\subsection{Decoupling Image-Based Input Features from Image-Based Pseudo-Ground-Truth Labels}
\label{sec:decoupling}

The second main issue is how best to use features derived from RGB images as input to the 3D network.  Features extracted from RGB images are very useful for 3D semantic scene understanding since many object classes are more easily recognized by their appearance in color images than by their geometric shapes (e.g., crosswalks). We demonstrate this quantitatively in the Appendix (+28.9 mIoU for our semantic 2D features, Table~\ref{tab:features}).

It would be natural to backproject semantic segmentation features extracted from 2D images, fuse them on 3D points and provide them as input channels for the 3D model.  But if this process matches the process used to create the pseudo-ground-truth labels, the input features and supervision will always agree at sparse supervision locations.
A network trained with these input features can minimize its loss by simply passing through the input features without doing any geometric reasoning.  
Such a pass-through function is not useful outside the sparse pseudo-ground truth set.
\footnote{Note that this problem is not encountered when 3D labels are available.  It occurs when the same set of image-based semantic segmentations are used to create both the input features and the ground truth.}

\begin{figure}
\centering
\newlength{\offsetpage}
\setlength{\offsetpage}{1.0cm}
\newenvironment{widepage}{\begin{adjustwidth}{-\offsetpage}{-\offsetpage}%
    \addtolength{\textwidth}{2\offsetpage}}%
{\end{adjustwidth}}
\includegraphics[width=0.4505\textwidth]{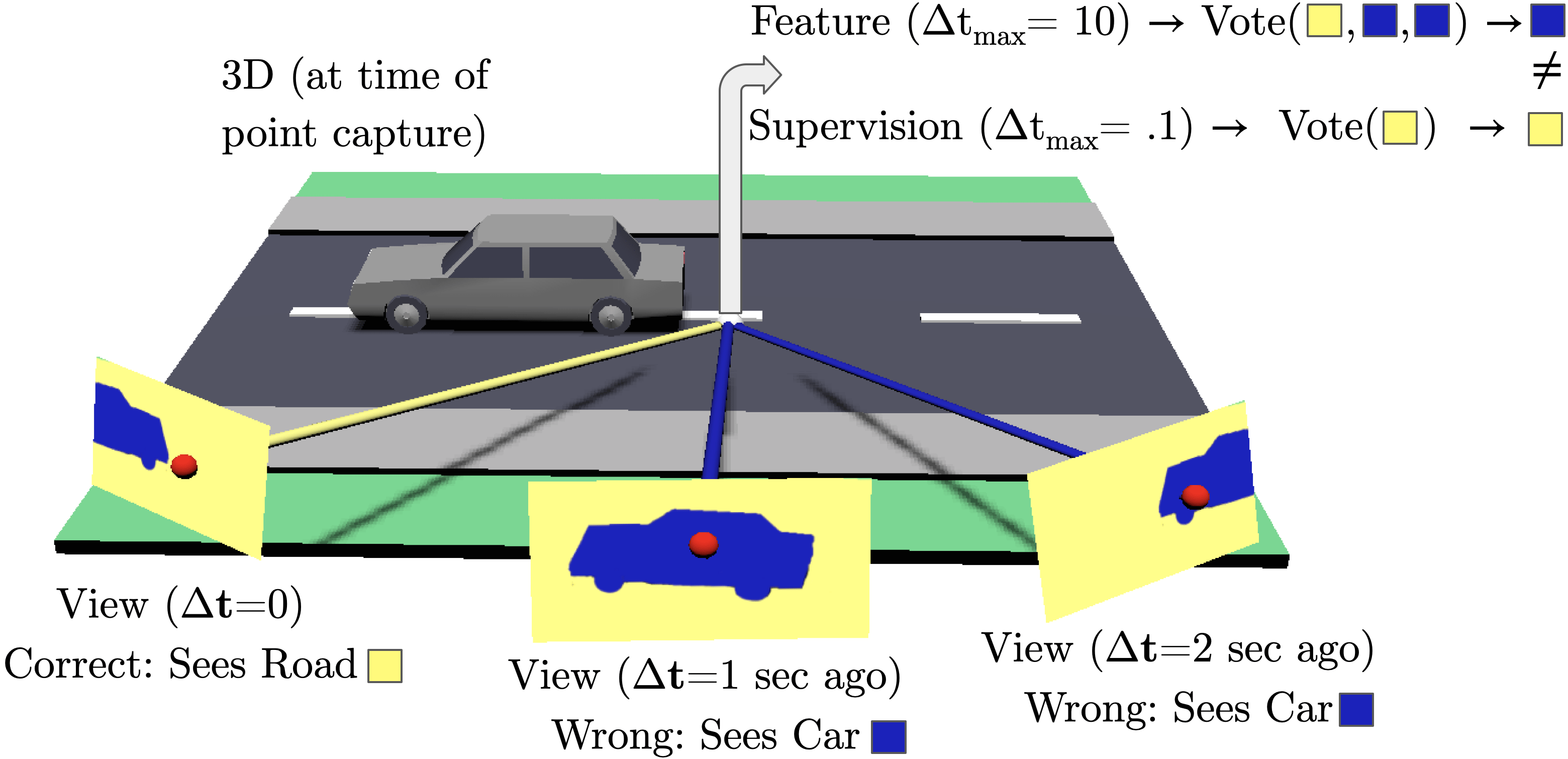}
\caption{\small{Illustration of how input features and pseudo-ground truth are generated for a point on the road.  Input features are created by voting with multiview fusion using all images -- in this case 2 out of 3 images vote ``car.''   Pseudo-ground truth is created by voting only with images captured at nearly the same time (small $\Delta$t) and with a clear line of sight (not near silhouettes) -- in this case, only the left image qualifies and the result is ``road.''  The 3D network is trained to predict the pseudo-ground truth from the input features -- i.e., fix the mistakes of multi-view fusion.}}
\label{fig:decorrelation}
\vspace{-1.15em}
\end{figure}
To address this issue, we aggregate input features for the network and pseudo-ground-truth labels using two independent multiview fusion steps that have different parameters. Specifically, when doing fusion for features, we increase the max timestamp difference $\Delta t_{max}$ substantially from $0.1$s to $10$s, reduce the dilation factor from $k=8$ to $k=2$, and relax the occlusion threshold from $\tau=1\%$ to $\tau=5\%$.

These parameter changes add more votes that exhibit many of the mistakes from Figure~\ref{fig:correspondenceproblem}, while also ensuring most points have a label. Critically, we could improve the quality of these features by removing some of the erroneous votes, but find it \textit{improves} performance to use decoupled features despite their lower accuracy (Section~\ref{sec:ablations}).
Decoupling teaches the network to correct mistakes, rather than pass through features. The result is a more useful function based on geometric reasoning. See Figure~\ref{fig:decorrelation} for an example.

An additional consideration is how to use 2D image information to create input features for the 3D network. Because supervision is only available at certain points, it is important that the 3D convolutions learned on supervised points transfer to unsupervised points. 
We find, for example, that RGB features can actually damage performance because they do not exist at points unobserved by any color image. When we transfer RGB features to unobserved points by copying from nearest neighbors, they still look unrealistic and unlike the training set, so the network performs badly in those regions.
Our choice is to aggregate only the final class label, encoded as a one-hot vector. This label is informative and compact enough to store. One-hot encodings transferred via nearest neighbor are more robust than RGB, because they are constant for the whole object. Most importantly, this label makes it difficult for the network to distinguish the pseudo-ground-truth points from other points.  This property makes convolutions learned at supervised points generalize well and makes class labels an ideal choice for our setting.


\subsection{Sampling Diverse Scenes}
\label{sec:sampling}

The third issue to address is how to sample training scenes from a large repository of data. In our system, a training scene is a short interval (10s) of data that typically includes a few million lidar points and a few dozen images.  The question is which scenes should we use for training when the repository is orders of magnitude larger than our practical processing capacity? 

The obvious answer is to sample scenes randomly.   However, then we might get a training set filled only with roads and trees.   It would be very unlikely to include rare objects, like phone booths or animals.  Accordingly, our trained models would be biased towards the common classes.

There is a long tail of complex and rare scenes, and we would like to find and train on those instead.  To do so, we use the pretrained 2D semantic segmentation network to estimate the objects in a scene, and then solve an optimization problem to select scenes containing images that provide a diverse training set.   Specifically, we first randomly select a huge set of candidate images $\mathcal{I}$ and run the 2D network on them.  For each image $I$, we check whether its semantic segmentation contains at least $p_{min}$ pixels for each of the $|C|$ class labels, forming a binary vector $v_{I} \in \{0, 1\}^{|C|}$.  We then create a training set $S$ of exactly $N$ images by maximizing the following (negative energy) objective function:

$$ 
\max_{S \subset \mathcal{I}} \quad  \norm{ \sum_{I \in \mathcal{S}} v_I}_{\frac{1}{2}}^{\frac{1}{2}}
\hspace{10mm}
\textrm{s.t.} \ |S| = N
$$ 

This optimization aims to find as many examples of every class as possible, with quadratically diminishing returns per class.  The set of scenes extracted from these images balances choosing individual classes that are rare with complex scenes that contain many objects. Note this approach is related to active learning~\cite{settles2009active, BUDD2021102062, NIPS2017_8ca8da41, aghdam2019active, haussmann2020scalable}, which has been used for segmentation datasets~\cite{8659293, vezhnevets2012active, Xie_2020_ACCV, Siddiqui_2020_CVPR, iglesias2011combining}. However, our case is distinct since our constraints come from processing an input example, not annotating it.


\begin{figure}[t]
    \centering
    \includegraphics[width=\columnwidth]{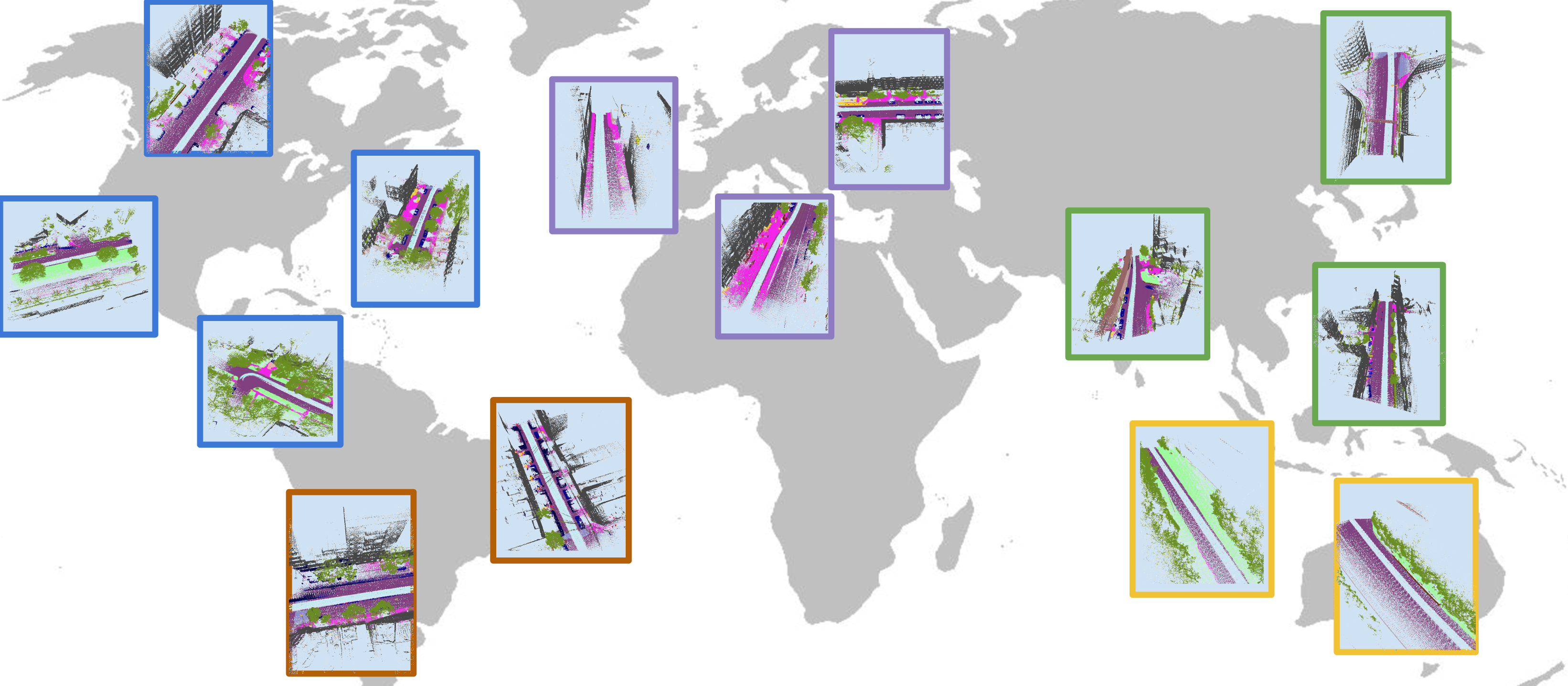}
    \caption{Our training and validation datasets include 3D scenes from cities around the world, a sampling of which are shown here.}
    \label{fig:worldmap}
    \vspace{-4mm}
\end{figure}

\section{Dataset}
\label{sec:dataset}

To evaluate the proposed algorithms in our target setting, we curated a dataset containing lidar and posed images collected from car-mounted scanners in 20 cities spread across 15 countries on 5 continents (Figure \ref{fig:worldmap}).   The scanning platform has 2 VLP16 lidar scanners mounted at oblique angles ($\sim45^{\circ}$) to scan building tops as well as street-level objects, and it has 7 color cameras with six in a hexagonal pattern and 1 pointing up (green lines in Figure \ref{fig:teaser}).   

Each of the 20 cities is divided into disjoint, interleaved sets of 2km x 2km square regions, one set for training and another for validation.  ``Scenes'' are sampled from these regions containing approximately ten seconds of data collection spanning approximately 1600 square meters with 3M lidar points and 60 color images on average.

The only training data available is a {\em previously labeled} set of 19,147 images captured in the training regions of the 20 cities along with 30 others.  Those images were manually annotated with 44 categories representing different types of ground (road, sidewalk, driveway, crosswalk, etc.), vehicles (car, truck, bicycle, etc.), urban structures (building, bridge, fence, etc.), street-side objects (traffic signs, streetlights, etc.), and person. See Appendix~\ref{sec:full_baseline_comparisons} for a full list.

To produce a validation set for 3D semantic segmentation, we selected one scene randomly from each of the 20 cities and employed an interactive 3D labeling tool to identify one of the 44 semantic categories for every lidar point in the scene.   This process took us several hours per scene using a highly optimized tool.   For budgetary reasons, it was not practical for us to label a large set of training scenes with the same process, since it would cost millions of dollars.

When testing algorithms, we evaluate predicted 3D semantic segmentations on the validation set with mean class IoU (mIoU) following standard protocols \cite{dai2017scannet}.
Since the validation set has limited data, we are careful when computing aggregate mIoU statistics to avoid averaging in results of classes with very little data.  Accordingly, we omit categories with less than 1000 points in each of at least 3 scenes when computing mean class IoU statistics.   
The remaining 26 classes include most with high interest to semantic mapping applications (e.g., building, road, sidewalk, crosswalk, car, truck, etc.) and 99.34\% of all scanned points.


\ignore{
\begin{table*}
\begin{center}
\setlength\tabcolsep{3pt} %
\begin{tabular}{l|cc|cccc|ccc}
\toprule
Method & mIoU $(\uparrow)$ & Mover mIoU $(\uparrow)$ & Road & Tree & Building & Sidewalk & Car & Person & Bicycle\\
\midrule
MVFusion & 48.1 & 43.0 & 84.8 & 88.8 & 91.6 & 65.3 & 46.6 & 19.9 & 54.7\\ 
Dense Supervision & 53.3 & 52.2 & 87.5 & 90.5 & 92.7 & 68.6 & 57.2 & 30.9 & 60.6\\ 
Ours & \textbf{59.5} & \textbf{68.9} & \textbf{91.5} & \textbf{92.5} & \textbf{94.3} & \textbf{72.6} & \textbf{83.1} & \textbf{48.8} & \textbf{70.8}\\
\bottomrule
\end{tabular}
\caption{Comparison to baseline methods. We report the mean as well as performance on the most common mover and non-mover classes. See Table~\ref{tab:full_quant} for results on all classes.
}
\label{tab:quant}
\end{center}
\vspace{-4mm}
\end{table*}
}

\section{Experimental Results}
\label{sec:results}
Here, we run experiments to compare the proposed approach to baselines and to test the impact of each component of our approach. Unless otherwise specified, evaluations are performed on the 20 city dataset from Section \ref{sec:dataset}.


\begin{figure*}
    \includegraphics[width=\textwidth]{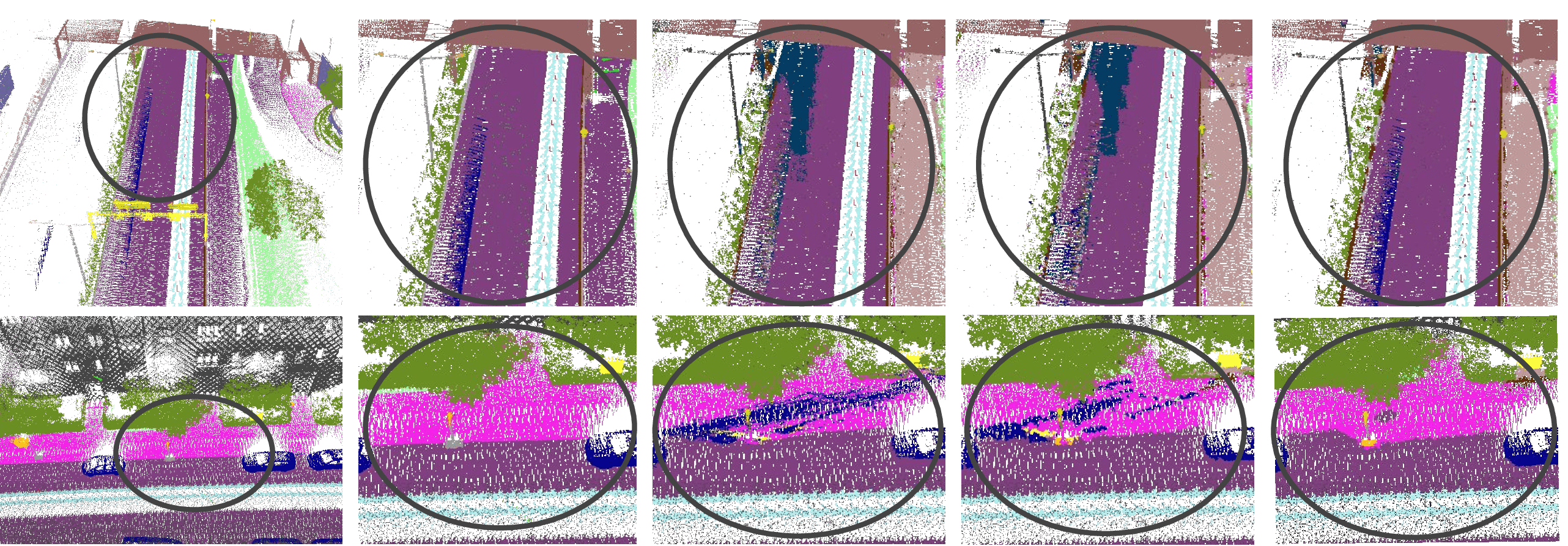}
    \hspace*{0.25em} Ground Truth (Context) \hspace{2.6em}{Ground Truth} \hspace{4.3em} MVFusion \hspace{3.35em} Dense Supervision \hspace{4.35em} Ours
    \caption{Qualitative results. Multiview fusion frequently makes nongeometric mistakes, especially for mover classes (votes for a bus end up on the ground, car and person on the sidewalk, building on person). Training a 3D network with dense supervision reduces these nongeometric mistakes slightly, but our approach improves the geometric prior substantially.}
    \label{fig:qual}
    \vspace{-4mm}
\end{figure*}

\subsection{Comparisons to 2D Supervised Baselines}
\label{sec:comparisons}

Our first set of experiments evaluates the performance of the proposed methods in comparison to alternatives that can work without direct 3D supervision.   
Since almost all prior methods are designed for benchmarks with 3D training sets, there are few baselines for our task.  To the best of our knowledge, multi-view fusion is the only/best alternative \cite{hermans2014dense}.   We compare to that and to a baseline we created to represent a straightforward implementation of our approach with dense rather than sparse pseudo-labels.   
\begin{itemize}
\setlength{\topsep}{0pt}
\setlength{\parsep}{0pt}
\setlength{\partopsep}{0pt}
\setlength{\parskip}{0pt}
\setlength{\itemsep}{2pt}
\vspace*{-2mm}
\item Multi-view fusion \cite{hermans2014dense} (MVF): This approach is a standard implementation of multi-view fusion that aggregates semantic labels at 3D points based on voting of labels at pixels with unoccluded sight lines.
\item Dense supervision (Dense): This approach performs multi-view fusion as described above and then trains a 3D network with dense supervision from the predicted labels.  Although this baseline is a variant of our main idea rather than previous work, it is valuable to investigate how our approach of making pseudo-ground-truth compares to a simpler alternative.
\end{itemize}

For both our method and these baselines, we train a DeepLab v3~\cite{chen2017rethinking} with Xception65~\cite{chollet2017xception} network on the labeled training images.   That network takes in an RGB image and outputs a dense image prediction with a class label per pixel, which is used for multi-view fusion by the baselines and pseudo-supervision for our 3D network.  For our 3D network, we use the SparseConv architecture~\cite{graham20183d}.

\begin{table}[t]
\begin{center}
\setlength\tabcolsep{2pt}
\begin{tabular}{l|cc|ccc|ccc}
\toprule
& Total & Mover & & & & & & \\
Method & mIoU & mIoU & Road & Tree & Bldg & Car & Person & Bike\\
\midrule
MVF & 48.1 & 43.0 & 84.8 & 88.8 & 91.6 & 46.6 & 19.9 & 54.7\\ 
Dense & 53.3 & 52.2 & 87.5 & 90.5 & 92.7 & 57.2 & 30.9 & 60.6\\ 
Ours & \textbf{59.5} & \textbf{68.9} & \textbf{91.5} & \textbf{92.5} & \textbf{94.3} & \textbf{83.1} & \textbf{48.8} & \textbf{70.8}\\ %
\bottomrule
\end{tabular}
\caption{Comparison to baseline methods. 
We report the mean as well as performance on the most common mover and non-mover classes. See Table~\ref{tab:full_quant} for results on all classes.
}
\label{tab:quant}
\end{center}
\vspace{-8mm}
\end{table}

\vspace*{2mm}\noindent{\bf Quantitative Comparisons}.  
Table~\ref{tab:quant} compares the quantitative evaluation metrics for our method and the baselines.   Overall, our method achieves 59.5 mIoU, while the best true baseline (MVFusion) achieves only 48.1.   Our method also outperforms the alternative proposed that trains a 3D network with supervision from dense multi-view fusion (53.3 mIoU). The largest improvements are for mover classes, with a jump of 25.9 mIoU over the MVFusion baseline. Our 3D network trained with sparse pseudo-supervision is able to better segment movers because it learns to discriminate classes based on their geometric shapes (e.g,. road vs.\ car).

\vspace*{2mm}\noindent{\bf Qualitative Comparisons}.  
Figure~\ref{fig:qual} shows examples of semantic segmentations produced by our method and baselines.  The other methods make errors that are clearly visible when looking at the 3D points (e.g. car labels on the ground and sidewalk). By contrast, our approach better accounts for both image and geometric features.

Typical errors for our method include 1) difficulty with classes that consistently differ between 2D and 3D, 2) missing small objects, and 3) failing to crisply separate 2D boundaries from badly fused features. 
For example, in Figure~\ref{fig:qual}, top, the road behind the fence is mislabeled as fence due to consistently incorrect supervision in the sparse pseudo-ground-truth, and is only slightly improved over standard MVFusion.  


\begin{table}[t]
\begin{center}
\setlength\tabcolsep{3pt} %
\begin{tabular}{l|cc}
\toprule
& Total & Mover \\
Method & mIoU $(\uparrow)$ & mIoU $(\uparrow)$\\
\midrule
Ours &  59.5 & 68.9\\
\midrule
No Diverse Scene Sampling & 49.2 & 32.0 \\ 
No 3D Convolutions & 52.5 & 50.0\\ 
No Sparse Supervision & 53.3 & 52.4\\ 
No Timestamp Filtering & 53.4 & 62.9\\ 
No Dilation & 59.8 & 68.5\\ 
\bottomrule
\end{tabular}
\caption{Method ablations. All rows show variants of our method with a single algorithmic component removed or altered. Diverse scene sampling, 3D convolutions, timestamp filtering, and sparse supervision are all critical to achieving good results.}
\label{tab:ablations}
\end{center}
\vspace{-8mm}
\end{table}

\subsection{Ablation Studies}
\label{sec:ablations}

Our second set of experiments investigates how each component of our system affects performance.  
Table~\ref{tab:ablations} shows results of a knock-out study, where each row shows results of our method with one algorithmic component disabled.  We find that diverse scene sampling, 3D convolutions, timestamp filtering, and sparse supervision are critical to achieving good results.  Diverse scene sampling improves performance the most, by 10.6 mIoU, even when training on 250K scenes and 750B points.  3D convolution provides the second most improvement, as the semantic features before 3D convolution have all the problems shown in Figure \ref{fig:correspondenceproblem}. Dilating occluders helps with some classes (fences, trees, bicycles), but has mixed results overall. 

In another experiment, we evaluate how import decoupling features is. We replace input features with pseudo-ground-truth values at pseudo-ground-truth locations. In other words, we increase coupling by \textit{improving} feature quality while holding supervision fixed. As expected, increased coupling reduces performance- this ablation has increased train (pseudo-ground-truth) performance (+8.8 mIoU) but reduced test (true) performance (-2.6 mIoU). See Appendix~\ref{sec:additional_ablations} for more details.

\subsection{Experiments with the nuScenes Benchmark}
\label{sec:nuscenes}
In this section, we evaluate our method using nuImages and the nuScenes lidarseg benchmark dataset~\cite{caesar2019nuscenes}. While the nuScenes setting is different than ours (it has in-domain 3D supervision and data in only 2 cities), it enables several important experiments. In particular, it allows us to 1) evaluate how well our proposed architecture compares to the state-of-the-art (SOTA), 2) evaluate how well our approach performs when no, partial or full 3D supervision is available, and 3) study the generalization benefits of our 2D pseudo-supervision approach.

\vspace*{2mm}\noindent\textbf{Is our 2D3DNet architecture SOTA?}
We first evaluate how our network architecture compares to previous work when given full 3D supervision.  We trained the 2D3DNet architecture on the nuScenes lidarseg dataset, using nuImages to train our 2D network for features, and submitted our test split results to the nuScenes leaderboard. Our score is 80.0 mIoU, which is 1st place (+1.7 mIoU) compared to all published methods, 3rd place overall including recent unpublished submissions, and 1st place among all 2D-3D fusion based approaches (+2.3 mIoU).  See Appendix~\ref{sec:nusc-sup} or the nuScenes lidarseg leaderboard for per-class mIoUs.   

\begin{figure}[t]
    \centering
    \vspace{-1.5em}
    \includegraphics[width=\columnwidth]{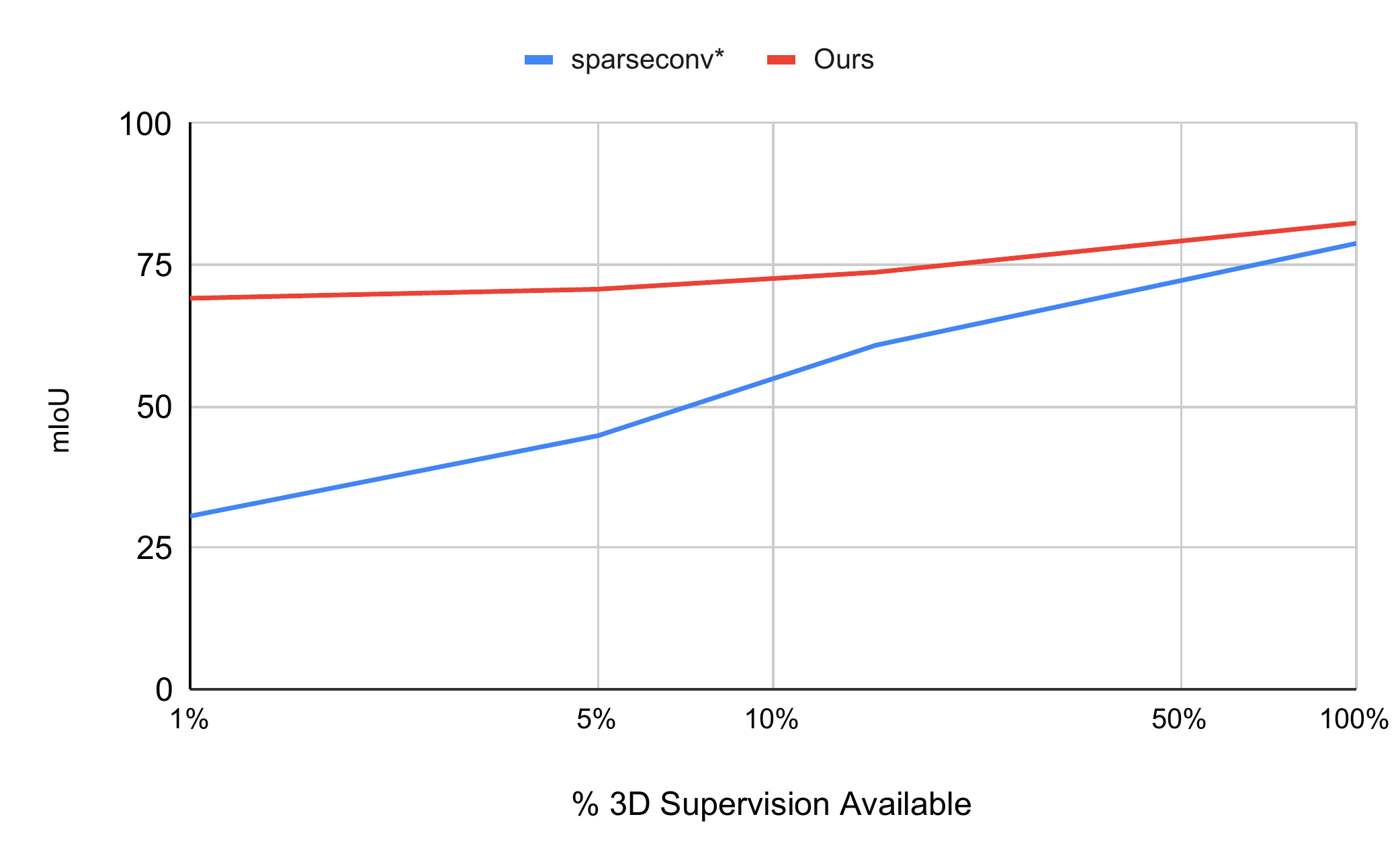}
    \vspace{-2.25em}
    \caption{Compared to standard 3D baselines, our approach learns more quickly from a small amount of 3D data.}
    \vspace{-1.25em}
    \label{fig:3dsup-quantity}
\end{figure}

\vspace*{2mm}\noindent\textbf{Is 2D supervision useful even when some 3D supervision is available?}
We next evaluate whether our pseudo-supervision and 2D features can improve performance when combined with varying amounts of 3D supervision. We compare two approaches. The first approach is 2D3DNet, where some fraction of training scenes use 3D annotated labels and all remaining scenes use pseudo-supervision. All pseudo-supervision is generated by a 2D model trained on nuImages. The second approach is the 3D portion of our architecture (``SparseConv*'') trained on 3D data. This SparseConv* baseline takes in surface normal features rather than our semantic features and is trained on 3D ground truth labels only; it is otherwise identical to our approach.
Figure \ref{fig:3dsup-quantity} provides mIoU results for several different sizes of the 3D training subset for both 2D3DNet (red) and SparseConv* (blue).   We evaluate on the 11 classes present in both nuScenes and nuImages. The difference between the two curves suggests that 2D3DNet is not as dependent on a large 3D training set. At 0\% supervision, 2D3DNet achieves 65.6 mIoU, versus 80.7 mIoU at 100\% supervision (92.0 vs 96.4 fwmIoU). By comparison, at 100\% 3D supervision SparseConv* reaches 78.8 mIoU, but only 30.6 mIoU at 1\%-- 2D3DNet's mIoU is 69.1 at 1\%.  These results suggest that 3D supervision is superior when it is available in sufficient quantity, but that our methods for utilizing labeled 2D image collections can help 3D networks achieve competitive performance even when far less 3D supervision is available.

\vspace*{2mm}\noindent\textbf{Does 2D supervision improve generalization?}  Finally, we evaluate whether our 2D pseudo-supervision improves generalization to new cities. We begin by partitioning the nuScenes lidarseg dataset into two subsets-- Boston (467 scenes) and Singapore (383 scenes). We then consider three possible supervision strategies for training a 3D SparseConv* network. 
In the first case, we train using only 3D supervision from one city, and test generalization to the other city. In the second case, we train using 3D supervision from one city and also add pseudo-supervision from nuImages for the other city. In the final case, we train using no 3D supervision at all, and instead use our pseudo-supervision approach for both cities. The performance of each of the the three cases for each city are in Table~\ref{tab:generalization}.

\begin{table}[t]
\begin{center}
\setlength\tabcolsep{3pt} %
\begin{tabular}{cc|cc}
\toprule
\multicolumn{2}{c}{Supervision} & \multicolumn{2}{c}{Validation}\\
\midrule
3D & 2D & Same City & Other City\\
\midrule
Singapore & None & 70.3 & 38.9 \\
Singapore & Boston & 68.6 & \textbf{64.3} \\
None & Both & 63.1 & \textbf{64.3} \\
\midrule
Boston & None & 76.1 & 55.6\\
Boston & Singapore & 74.8 & \textbf{68.8}\\
None & Both & 64.3 & 63.1 \\
\bottomrule
\end{tabular}
\vspace{-0.5em}
\caption{Adding our 2D pseudo-supervision helps SparseConv generalize better to cities for which there is no available 3D training data. The top three rows use Singapore as the 3D supervised city, and the bottom three use Boston. Rows 3 and 6 are the same model, since no 3D supervision is used in either case.}
\label{tab:generalization}
\end{center}
\vspace{-8mm}
\end{table}

We observe that: 1) Models trained only on 3D supervision from one city do not generalize well to the other city (38.9 Singapore$\rightarrow$Boston, 55.6 Boston$\rightarrow$Singapore). 2) 2D pseudo-supervision from both cities outperforms 3D supervision from a different city (64.3 vs 38.9 Boston, 63.1 vs 55.6 Singapore). 3) Adding 3D supervision from a different city on top of 2D pseudo-supervision improves performance only slightly (64.3 vs 64.3 Boston, 68.8 vs 63.1 Singapore). These observations suggest that our approach is a good alternative for 3D semantic segmentation in cities around the world for which 3D labeled data is not available.

\section{Conclusion}
\label{sec:conclusion}

In this paper, we investigate how to learn 3D semantic segmentation without any 3D labeled training data.  We propose creating trusted pseudo-labels, leveraging 2D semantic predictions as input features to a 3D network, and efficiently sampling diverse scenes from a large unlabeled repository.  These methods enable training 3D semantic segmentation models that are 6.2-11.4 mIoU better than baselines on a validation dataset sampled from 20 cities on five continents, and 16.7-25.9 mIoU better on mover classes. Further experiments with the nuScenes lidarseg dataset show that 2D3DNet outperforms published methods (+1.7 mIoU) when given full supervision, our architecture is more robust than 3D supervised baselines when only a small amount of supervision is available (+12.9-38.5 mIoU), and that pseudo-supervision helps with generalization to novel cities (+7.5-25.4 mIoU). These results suggest promising avenues for leveraging repositories of image data to train generalizable 3D models that can be deployed at scale.

\appendix
\section*{Appendix}

This appendix contains additional information in support of the main paper.   Specifically, it includes:
\vspace{-0.4\topsep}
\begin{itemize}
\setlength{\itemsep}{2pt}
\setlength{\parskip}{2pt}

\item \textbf{Expanded supervised nuScenes results} reporting the quantitative performance of 3D supervised models on the test and validation splits of the lidarseg task (Appendix \ref{sec:nusc-sup}).

\item \textbf{Additional ablation studies} investigating the impact of the key algorithms of the proposed method in more detail (Appendix \ref{sec:additional_ablations}).

\item \textbf{Comparison to a SemanticKITTI 3D supervised model} demonstrating the difficulty of transferring 3D supervision between datasets (Appendix \ref{sec:kitti_comparison}).

\item \textbf{Additional experiments on ScanNet} investigating how the proposed approach can work without any in-domain supervision at all (Appendix \ref{sec:additional_scannet}).

\item \textbf{Additional baseline comparison statistics} providing results for all 44 classes of the twenty city dataset (Appendix \ref{sec:full_baseline_comparisons}).

\item \textbf{Additional qualitative visualizations} showing the output of our model and fused semantic features (Appendix \ref{sec:full_qual}).

\item \textbf{Implementation details} for the 2D and 3D network architectures, multiview fusion, normal estimation, decoupling, optimization algorithm, and evaluation metrics (Appendix \ref{sec:implementation}).
\end{itemize}

\section{nuScenes Supervised Results}
\label{sec:nusc-sup}
In Table~\ref{tab:nuscenes-test}, we show the performance of our method on the test set (that is, we show the numbers available at \texttt{www.nuscenes.org/lidar-segmentation} on the nuScenes leaderboard). For all 16 classes, please see the leaderboard. To facilitate comparisons, we also provide our validation split numbers in Table~\ref{tab:nuscenes-val} (no comparison to SPVCNN++ or GU-Net is possible because the details and code for those methods are unavailable, though we add a comparison to the SparseConv* architecture used for some experiments in Section~\ref{sec:nuscenes}). Our result is the highest published method on the test set, the 3rd highest overall (including recent unpublished submissions), and to our knowledge the highest reported on the validation split. 

\begin{table}[t]
\begin{center}
\setlength\tabcolsep{2.25pt}
\begin{tabular}{l|c|ccc}
\toprule
Method & mIoU & Pedestrian & Car & Bicycle\\
\midrule
SPVNAS~\cite{tang2020searching} & 77.4 & 75.6 & \textbf{90.8} & 30.0\\
Cylinder3D++~\cite{zhu2020cylindrical} & 77.9 & 77.3 &  89.4 & 33.9\\
AF2S3Net~\cite{cheng20212} & 78.3 & 77.3 & 84.2 & 52.2\\
\textcolor{gray}{GU-Net} & \textcolor{gray}{80.3} & \textcolor{gray}{81.3} & \textcolor{gray}{91.4} & \textcolor{gray}{33.4}\\
\textcolor{gray}{SPVCNN++} & \textcolor{gray}{81.1} & \textcolor{gray}{83.4} & \textcolor{gray}{92.2} & \textcolor{gray}{43.1}\\
\midrule
\textcolor{gray}{LIFusion} & \textcolor{gray}{75.7} & \textcolor{gray}{80.3} & \textcolor{gray}{84.3} & \textcolor{gray}{36.3}\\
\textcolor{gray}{CPFusion} & \textcolor{gray}{77.7} & \textcolor{gray}{78.1} & \textcolor{gray}{86.2} & \textcolor{gray}{37.0}\\
Ours & \textbf{80.0} & \textbf{82.0} & 85.1 & \textbf{59.4}\\
\bottomrule
\end{tabular}
\vspace{0.1em}
\caption{Results on the test split of the nuScenes lidarseg task~\cite{caesar2019nuscenes}. Methods above the bar are lidar only; methods below the bar use both 2D and 3D reasoning. Methods in gray are currently unpublished and have no details besides the leaderboard description. We bold the highest published result. For up to date results on all 16 classes see the nuscenes leaderboard at \texttt{nuscenes.org/lidar-segmentation} (our entry is anonymized).
}
\label{tab:nuscenes-test}
\end{center}
\vspace{-8mm}
\end{table}

\begin{table*}[t]
\begin{center}
\setlength\tabcolsep{2.25pt}
\begin{tabular}{l|c|cccccccccccccccc}
\toprule
Method & mIoU & \rotatebox{70}{Barrier} & \rotatebox{70}{Bicycle} & \rotatebox{70}{Bus} & \rotatebox{70}{Car} & \rotatebox{70}{Construction} & \rotatebox{70}{Motorcycle} & \rotatebox{70}{Pedestrian} & \rotatebox{70}{Cone} & \rotatebox{70}{Trailer} & \rotatebox{70}{Truck} & \rotatebox{70}{Driveable} & \rotatebox{70}{Other Flat} & \rotatebox{70}{Sidewalk} & \rotatebox{70}{Terrain} & \rotatebox{70}{Manmade} & \rotatebox{70}{Vegetation}\\
\midrule
RangeNet++~\cite{milioto2019iros} & 65.5 & 66.0 & 21.3 & 77.2 & 80.9 & 30.2 & 66.8 & 69.6 & 52.1 & 54.2 & 72.3 & 94.1 & 66.6 & 63.5 & 70.1 & 83.1 & 79.8\\
PolarNet~\cite{zhang2020polarnet} & 71.0 & 74.7 & 28.2 & 85.3 & 90.9 & 35.1 & 77.5 & 71.3 & 58.8 & 57.4 & 76.1 & 96.5 & 71.1 & 74.7 & 74.0 & 87.3 & 85.7\\
Salsanext~\cite{cortinhal2020salsanext} & 72.2 & 74.8 & 34.1 & 85.9 & 88.4 & 42.2 & 72.4 & 72.2 & 63.1 & 61.3 & 76.5 & 96.0 & 70.8 & 71.2 & 71.5 & 86.7 & 84.4\\
SparseConv* & 75.2 & 76.9 & 25.4 & 91.5 & 86.8 & 54.0 & 76.7 & 77.9 & 62.0 & 66.4 & 83.0 & 96.0 & 74.7 & 75.9 & 75.2 & 91.3 & \textbf{89.9}\\
Cylinder3D~\cite{zhu2020cylindrical} & 76.1 & 76.4 & 40.3 & 91.2 & \textbf{93.8} & 51.3 & 78.0 & 78.9 & 64.9 & 62.1 & 84.4 & \textbf{96.8} & 71.6 & \textbf{76.4} & \textbf{75.4} & 90.5 & 87.4\\
Ours & \textbf{79.0} & \textbf{78.3} & \textbf{55.1} & \textbf{95.4} & 87.7 & \textbf{59.4} & \textbf{79.3} & \textbf{80.7} & \textbf{70.2} & \textbf{68.2} & \textbf{86.6} & 96.1 & \textbf{74.9} & 75.7 & 75.1 & \textbf{91.4} & \textbf{89.9}\\
\bottomrule
\end{tabular}
\vspace{-0.5em}
\caption{Results on the validation split of the nuScenes lidarseg task~\cite{caesar2019nuscenes}.}
\label{tab:nuscenes-val}
\end{center}
\vspace{-8mm}
\end{table*}

\section{Additional Ablation Studies}
\label{sec:additional_ablations}

In this section, we report results of experiments investigating ablations of our method.



\vspace*{2mm}\noindent\textbf{How does feature decoupling affect performance?}
Table~\ref{tab:disagreement} demonstrates that feature decoupling improves performance. In this experiment, we use identical sparse supervision for both approaches. However, in the ablation (Abl), we improve the feature quality by using the pseudo-ground truth fusion where available. In other words, we ablate the decoupling so that the supervision remains the same but the feature quality improves. Despite more accurate features, the model performance is worse due to increased correlation between the features and the pseudo-ground truth.

\begin{table}[t]
\begin{center}
\setlength\tabcolsep{2pt}
\begin{tabular}{l|cc|cc}
\toprule
& \multicolumn{2}{c}{All} & \multicolumn{2}{c}{Movers}\\
Method & Train & Test & Train & Test\\

\midrule
Ours & 61.5 & \textbf{60.1} & 59.1 & \textbf{66.1}\\
Abl. & 70.3 & 57.5 & 72.9 & 61.0\\
\bottomrule
\end{tabular}
\vspace{-0.5em}
\caption{Ablation of feature decorrelation. Train (pseudo-ground truth) mIoU increases and test (true) mIoU decreases.
}
\label{tab:disagreement}
\end{center}
\vspace{-6mm}
\end{table}

\vspace*{2mm}\noindent\textbf{What is the effect of 2D features?}  
Table \ref{tab:features} shows results of a study investigating the performance of different features provided with 3D points as input to the 3D network.  As expected, we find that it is critical to utilize the proposed semantic features derived from the 2D images (one-hot encoding of the semantic class generated by fusing predictions of the 2D network).  Counter-intuitively, adding RGB channels as an additional feature hurts performance (by 1.7\% mIoU).  We conjecture the reason is due to a test-time domain shift for lidar points not observed by any RGB image.  For those points, we make our best effort to fill in missing RGB features by copying from nearest spatial neighbors.  However, the resulting RGB features are out of domain for the pseudo-ground-truth points used for training, which are all observed directly by RGB images, leading to an adverse affect on performance.   Yet, interestingly, the same is not true when using one-hot encodings of semantic labels.   Since the network is trained with a protocol that decouples input semantic features from pseudo-ground-truth labels (Section~\ref{sec:decoupling}), it learns to compensate for wrong semantic input labels and generalizes better to all points.

\begin{table}[t]
\begin{center}
\setlength\tabcolsep{3pt} %
\begin{tabular}{l|cc}
\toprule
Features & mIoU $(\uparrow)$ & Mover mIoU $(\uparrow)$\\
\midrule
Semantic & \textbf{59.5} & \textbf{68.9}\\
Semantic + RGB & 57.8 & 65.8\\ 
Normals & 27.7 & 21.8\\ 
Normals + RGB & 30.6 & 26.9\\ 
\bottomrule
\end{tabular}
\caption{Impact of different input features to the 3D network. One-hot encoding of the semantic class predicted by the 2D network is best. Semantic + Normals was also investigated but found to be unhelpful.}
\label{tab:features}
\end{center}
\vspace{-1.5em}
\end{table}


\vspace*{2mm}\noindent\textbf{How effective is the diverse scene sampling?}
\label{sec:scaling_experiments}
Table~\ref{tab:dss} shows more details on how the diverse scene sampling algorithm affects performance.  For this study, we test training datasets with 250, 2.5K, 25K, and 250K scenes, created both with and without our diverse scene sampling strategy (Section~\ref{sec:sampling}).   We find that diverse sampling provides a significant performance improvement over random sampling at all dataset sizes (10.3\% mIoU on the largest dataset).   The set of test scenes is sufficiently complex and varied that diverse training data improves generalizability and performance on multiple classes (other vehicle, motorcycle, street light, bicycle in particular).  In contrast, the results achieved with a very large training dataset of random scenes (250K scenes) are not better than with a very small dataset (250 random scenes).  We conjecture that it would require an impractically large training set of random scenes and a new loss carefully designed to handle extreme class imbalance to achieve comparable results to our diverse scene sampling strategy.

\begin{table}[t]
    \begin{center}
    \begin{tabular}{c|cccc}
    \toprule
    \multicolumn{1}{c}{Method} & \multicolumn{4}{c}{mIoU}\\
    \midrule
    \# Scenes & 250 & 2.5K & 25K & 250K\\
    \midrule
    Random Sampling & 53.1 & 50.8 & 51.3 & 49.2\\
    Diverse Sampling & 58.1 & 58.3 & 58.7 & 59.5\\
    \bottomrule
    \end{tabular}
    \caption{Our approach makes efficient use of large unsupervised data sets. By contrast, random sampling performs worse at all dataset sizes and does not reliably improve performance as the dataset grows (rare cases remain rare and difficult to learn).} 
    \label{tab:dss}
    \end{center}
    \vspace{-8mm}
\end{table}


\section{SemanticKITTI Supervision Transfer}
\label{sec:kitti_comparison}

In this experiment, we explore an alternative to 2D supervision-- transferring the 3D supervision from an existing dataset. This alternate approach avoids many of the challenges discussed in Section~\ref{sec:pseudolabels}. It also requires generalizing to a new terrestrial platform and to unseen cities, countries, and continents. 

Table~\ref{tab:alt-supervision} shows a comparison to a state of the art model, Cylinder3D~\cite{zhu2020cylindrical}, trained with 3D supervision from SemanticKITTI~\cite{behley2019semantickitti} and tested on our twenty city dataset.  We choose SemanticKITTI because of its class overlap with our evaluation set. To adapt our input scans as directly as possible, we provide the pair of full LiDAR sweeps as inputs to their network for each scan.  To provide a fair evaluation, we consider only classes shared between the two evaluation sets, and we report only the IoU of classes with the best performance for~\cite{zhu2020cylindrical} in Table~\ref{tab:alt-supervision}. 
Not surprisingly, this simple baseline is not competitive with our approach due to differences in the sensor patterns (different lidar angles) and scene content (1 training city versus 20 evaluation cities).   Yet, the large performance gap (48.7 points) highlights the problems of domain adaptation on 3D point clouds.

\begin{table}
\begin{center}
\setlength\tabcolsep{2.25pt}
\begin{tabular}{c|c|ccccccc}
\toprule
  & Mean & Bld. & Road & Car & SW. & Fence & Pole & Bike \\
\midrule
Ours & \textbf{72.7} & \textbf{96.1} & \textbf{93.3} & \textbf{84.0} &\textbf{77.8} & \textbf{60.7} & \textbf{60.7} & \textbf{75.8} \\
\cite{behley2019semantickitti,zhu2020cylindrical} & 24.0 & 76.3 & 38.9 & 68.5 & 28.4 & 18.2 & 12.9 & 10.1 \\
\bottomrule
\end{tabular}
\vspace{0.1em}
\caption{A comparison to a state of the art 3D model (Cylinder3D~\cite{zhu2020cylindrical}) trained with 3D supervision on SemanticKITTI~\cite{behley2019semantickitti}. 
}
\label{tab:alt-supervision}
\end{center}
\vspace{-8mm}
\end{table}

\section{Application to Indoor Datasets (ScanNet)}
\label{sec:additional_scannet}

\begin{figure}[t]
    \centering
    \includegraphics[width=0.45\textwidth]{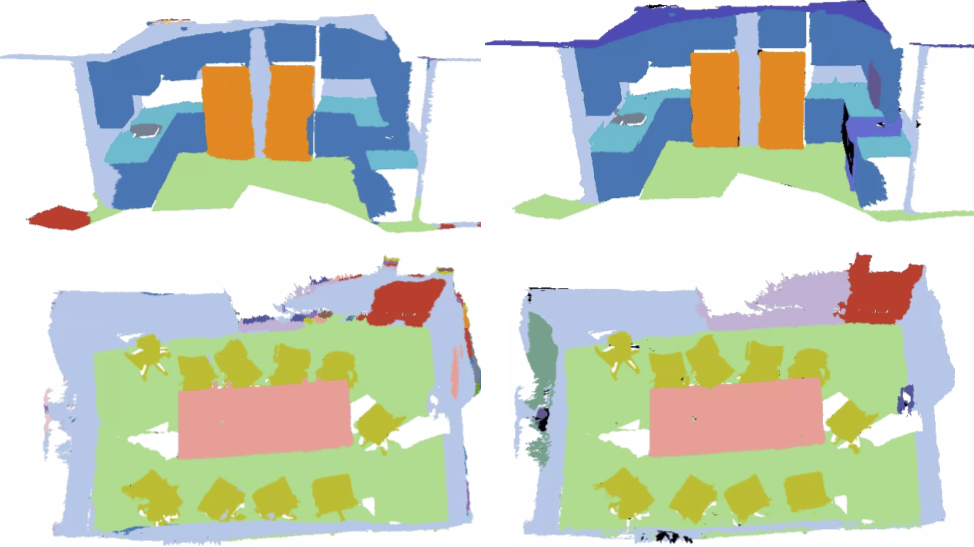}
    \caption{ScanNet 3D segmentations using our network  trained only from MSeg images (no ScanNet data at all) and tested only with 3D point inputs (no images from the test scenes).  The 3D predictions (left) are originally for the 194 MSeg classes; we map onto the 20 ScanNet classes for display purposes to compare to the ground truth (right).}
    \label{fig:scannet}
    \vspace{-5mm}
\end{figure}

As an experiment to test the generality of our main ideas under extreme settings, we ran an experiment on the ScanNet Semantic Segmentation Benchmark \cite{dai2017scannet} using only 3D point cloud inputs and no ground truth data from ScanNet at all (no 3D point labels, no image labels, and not even the ScanNet category list).  To train our 3D network, we used pseudo-supervision derived from a pretrained MSeg Universal 2D model~\cite{lambert2020mseg} (which was trained on several 2D datasets, not including ScanNet).  Specifically, we predicted semantic segmentations of images from the ScanNet training set using the MSeg model, ran our algorithms to create sparse pseudo-ground-truth from those segmentations, trained our 3D model, and tested it on inputs that include only 3D points (using 3D normals as input features), and then evaluate the result using an approximate mapping between MSeg and ScanNet classes.  The result is 42.6 mIoU on the ScanNet validation set.   Example results are shown in Figure~\ref{fig:scannet}.  Of course, these results are not as good as methods trained with supervision from either 2D or 3D ScanNet data.   However, it is remarkable that it can perform so well (9 points better than PointNet++~\cite{qi2017pointnet++} trained with full 3D supervision from ScanNet) when the input at inference is only 3D geometry, and the supervision at training comes only from labeled 2D image collections from other domains, in spite of an ambiguous mapping from MSeg to ScanNet classes.

\section{Expanded Baseline Comparison Statistics}
\label{sec:full_baseline_comparisons}

\begin{table*}[t]
    \centering
    \begin{tabular}{l|ccc|cc}
    \toprule
Class & MVFusion & Dense Supervision & Ours & \# GT Points & \# GT Scenes\\
\midrule
SELF & 97.5 & 97.6 & 98.7 & 4,135,276 & 20\\
BUILDING & 91.6 & 92.7 & 94.3 & 18,430,607 & 17\\
TREE & 88.8 & 90.5 & 92.5 & 12,976,114 & 19\\
PAVED\_ROAD & 84.8 & 87.5 & 91.5 & 13,030,871 & 20\\
SIDEWALK & 65.3 & 68.6 & 72.6 & 4,356,483 & 18\\
CROSSWALK & 48.9 & 56 & 61.8 & 317,971 & 4\\
DRIVEWAY & 23 & 28.4 & 25.2 & 333,199 & 6\\
TERRAIN & 70.6 & 72.3 & 75.6 & 3,489,174 & 16\\
OTHER\_STRUCTURE & 35 & 39.5 & 42.9 & 373,576 & 12\\
FENCE & 42.1 & 45.5 & 47.2 & 818,815 & 15\\
WALL & 25.6 & 32.7 & 33.5 & 384,525 & 9\\
GUARD\_RAIL & 31.2 & 33.9 & 33.9 & 113,600 & 3\\
TRAFFIC\_SIGN & 43.2 & 49.7 & 58.7 & 56,779 & 10\\
BRIDGE & 88.1 & 90 & 90.7 & 923,639 & 3\\
STREET\_LIGHT & 50.2 & 59.3 & 66.8 & 14,355 & 4\\
BUSINESS\_SIGN & 9.2 & 23.5 & 27.3 & 26,630 & 3\\
OTHER\_PERMANENT\_OBJECT & 21.9 & 19.7 & 20.5 & 77,679 & 12\\
OTHER\_TEMP\_OBJECT & 40.1 & 43.3 & 48 & 353,521 & 15\\
CAR & 46.6 & 57.2 & 83.1 & 1,543,843 & 19\\
TEMP\_TRAFFIC\_SIGN & 2.7 & 1.4 & 0 & 6,836 & 3\\
TRUCK & 57.9 & 63.8 & 78.9 & 1,898,201 & 10\\
MOTORCYCLE & 32.1 & 38.8 & 51.2 & 31,537 & 4\\
POLE & 31.4 & 39.1 & 51.7 & 152,212 & 20\\
PERSON & 19.9 & 30.9 & 48.8 & 123660 & 10\\
OTHER\_VEHICLE & 47.1 & 62.1 & 80.4 & 15014 & 3\\
BICYCLE & 54.7 & 60.6 & 70.8 & 72763 & 4\\
\midrule
SKY & - & - & - & 0 & 0\\
TUNNEL & - & 0 & 0 & 0 & 0\\
BUS\_STOP & 2.8 & 2.4 & 0 & 34,856 & 1\\
PHONE\_BOOTH & 0 & 0 & 0 & 3,349 & 1\\
UNPAVED\_ROAD & 0 & 0 & 0 & 0 & 0\\
TRAFFIC\_LIGHT & 31 & 36.8 & 53.4 & 2,972 & 1\\
OTHER\_GROUND & 0 & 0 & 0 & 554 & 0\\
PARKING\_METER & 0.1 & 0 & 0 & 4,039 & 1\\
MAILBOX & 20.3 & 13.5 & 0 & 2,659 & 1\\
FIRE\_HYDRANT & 50.7 & 4.6 & 0 & 2,112 & 1\\
BILLBOARD & 0 & 0 & 0 & 4,424 & 2\\
TEMP\_CONE & 18.1 & 25.4 & 0 & 4,554 & 1\\
BUS & 0 & 0 & 0 & 0 & 0\\
VEHICLE\_ON\_RAILS & 0.3 & 0 & 0 & 887 & 0\\
MOUNTAIN & 0 & 0 & - & 0 & 0\\
ANIMAL & 0 & 0 & 0 & 141 & 0\\
WATER & 0 & 0 & - & 0 & 0\\
RIDER & 0.2 & 0.3 & 5.8 & 30606 & 2\\
\bottomrule
    \end{tabular}
    \caption{Quantitative results on the full set of 44 classes. For each class, we also report the number of ground truth points and the number of ground truth scenes containing at least 1,000 points. The criterion for inclusion in the evaluation in Table~\ref{tab:quant} was at least 3 scenes with at least 1,000 points each (note some classes have tens of millions of points across 15+ scenes). Classes above the bar were included in evaluation.}
    \label{tab:full_quant}
\end{table*}

\begin{figure}[t]
    \centering
    \includegraphics[width=0.45\textwidth]{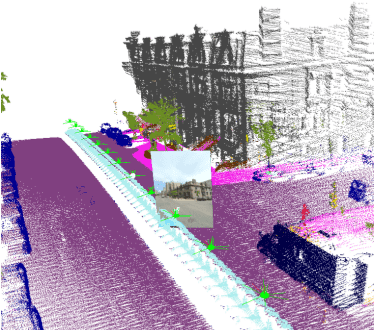}
    \caption{Our input ``scenes'' contain a sequence of RGB images from 7 cameras operating at $\sim$2Hz (camera extrinsics shown as bright green lines), and two independent lidar sensors capturing 16 beam sweeps at $\sim$10Hz. 
    }
    \label{fig:setting-explainer}
    \vspace{-4mm}
\end{figure}

Table~\ref{tab:full_quant} reports full quantitative statistics for the baseline comparison experiment described in Section~\ref{sec:comparisons} (i.e., it is an expanded version of Table~\ref{tab:quant}).  In addition to reporting per-class IoU for all 44 classes in the twenty city dataset, it also reports the numbers of points and GT scenes per class, which were used to determine the set of 26 classes for computing the mIoU in Table~\ref{tab:quant} (as described in Appendix \ref{sec:evaluation_details}). We also provide a visualization from one ground truth scene for illustrative purposes in Figure~\ref{fig:setting-explainer}.

\section{Additional Qualitative Results}
\label{sec:full_qual}
\begin{figure*}
    \centering
    \vspace{0.7em}
    \includegraphics[width=\textwidth]{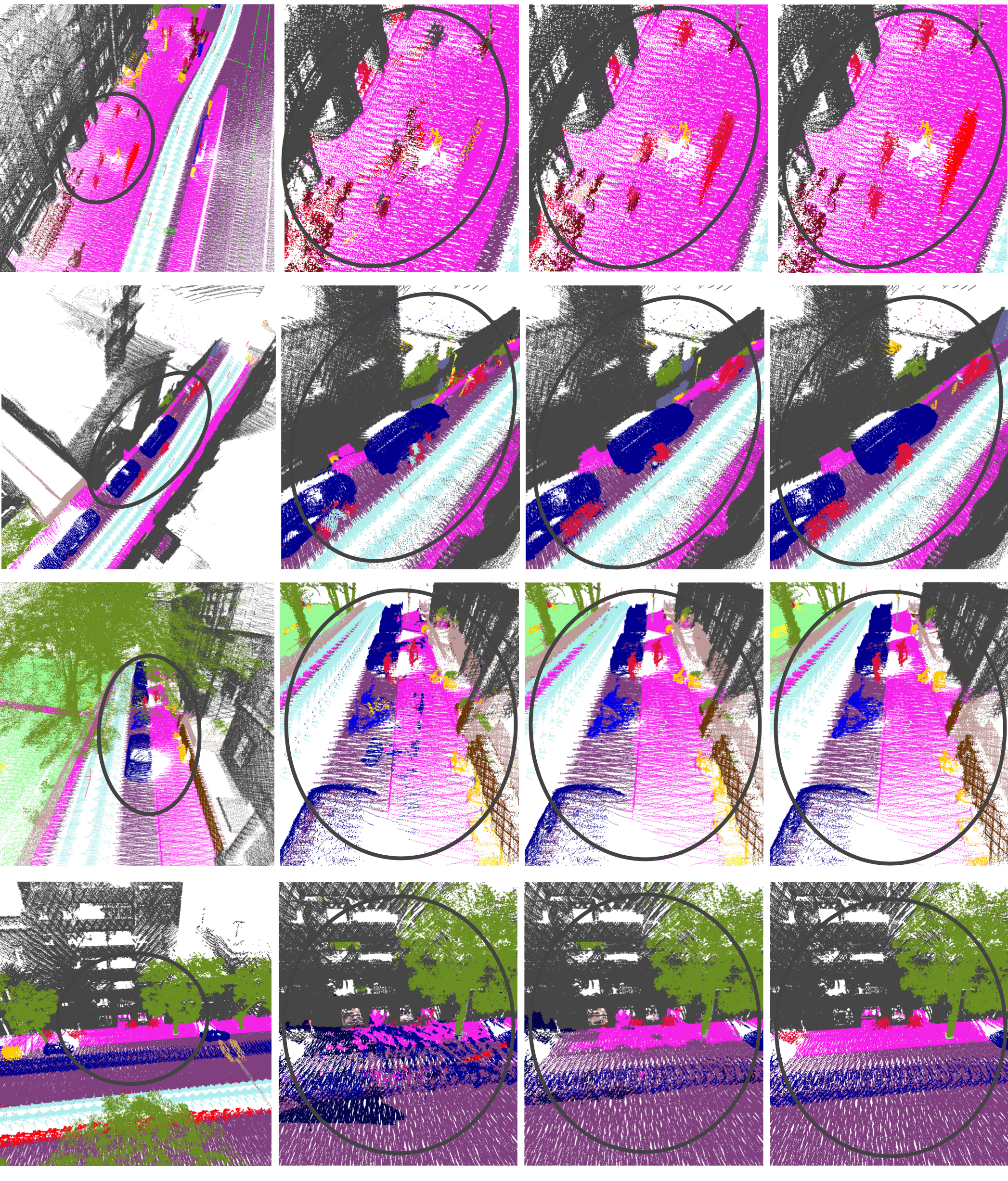}
     \hspace*{-0.625em}Ground Truth (Context) \hspace{3.75em} MVFusion (Alg.~\ref{alg:fusion}) \hspace{7em}{Ours} \hspace{8.25em}{Ground Truth}\\
     
    \caption{Qualitative visualizations on lidar scenes from our 20-city evaluation set. We visualize our first-stage dense label aggregations, the final output labels, and the ground truth predictions.}
    \label{fig:fullqual1}
\end{figure*}

In Figure~\ref{fig:fullqual1}, we provide additional qualitative visualizations of our method on our 20-city evaluation set. For each, we provide a ground truth visualization for context, as well as close-up comparisons of our intermediate multiview fusion predictions (as generated by Algorithm~\ref{alg:fusion} with the hyperparameters described in Section~\ref{sec:pseudolabels}), our final output predictions, and the ground truth.

\section{Implementation Details}
\label{sec:implementation}

In this section, we provide low level implementation details required for full reproducibility. We discuss network architecture and optimizer details, normal and radii estimation, multiview fusion, optimization details, and our evaluation procedure.

\subsection{Fusion Details}
The important aspects of our multiview fusion algorithm are described in Section~\ref{sec:pseudolabels}. We describe the lower level implementation details here with the goal of ensuring reproducibility. Most of the details that are omitted in Section~\ref{sec:pseudolabels} exist to improve runtime efficiency (e.g., the addition of a kdtree, the specific order of correspondence checks to avoid unncessary projections). There is also one additional parameter, $\Delta t_{thresh}$, which is set to 20 seconds in all cases. When a point is about to be rasterized into a depth image as a surfel, the point-image time difference is compared to $\Delta t_{thresh}$; points exceeding the threshold are not rasterized into the image. This parameter, too, exists primarily for performance reasons-- typically it only affects the resulting images when the car is still for long periods of time and there is extraneous data that does not need to be rendered to get a reasonable occlusion map ($\Delta t_{thresh}$ does this culling). All other parameters are described in Section~\ref{sec:pseudolabels}. Our algorithm is implemented in C++ and pseudocode is available in Algorithm~\ref{alg:fusion}.

Fusion parameters for our dataset are given in Section~\ref{sec:pseudolabels} ($d_{max} = 400$). For most nuScenes experiments, hyperparameters are largely unchanged, though we adjust the supervision $\Delta t_{max}$ to 0.085 to account for the different image sampling rate in nuScenes compared to our dataset (12 vs 2 hz). Before submitting to the leaderboard, we did additional experiments with 3D supervision to determine more optimal feature parameters when decoupling from 2D supervision is not a factor. We found that reducing $d$ to 1.5, increasing $\tau$ to $10\%$, and adding an additional term to $w_{ij}$
to scale inversely with class rarity (i.e., divide by the fraction of lidar points in the train split for that class) improved performance slightly, because they enable the network to better identify and correctly classify small and rare classes. These changes were made only for the leaderboard submission.

\begin{algorithm}
\For{int i = 0; i $<$ num\_points; i++}{
    Estimate normals and radii for point i\;
    Add point i to a kdtree for fast lookups\;
}
Initialize correspondences per point to an empty list\;
\For{int i = 0; i $<$ num\_images; i++}{
    Initialize empty depth channel for image i\;
    Create frustum from camera params and $d_{max}$\;
    Lookup points in frustum using kdtree\;
    \For{int j = 0; j $<$ num\_points\_in\_frustum; j++}{
        \If{point-image pair exceeds $\Delta t_{thresh}$}{
            continue\;
        }
        Use normal and radii to make disk for point\;
        Dilate the disk's radii by a factor of $d$\;
        Rasterize point $j$ as surfel to depth image $i$\;
    }
    \For{int j = 0; j $<$ num\_points\_in\_frustum; j++}{
    
        \If{point-image pair exceeds $\Delta t_{max}$}{
            continue\;
        }
        \If{point is backfacing to camera}{
            continue\;
        }
        Project point into image\;
        \If{point is outside image (rare)}{
            continue\;
        }
        Compute projected depth of point\;
        Lookup depth in depth image\;
        \If{relative depth error exceeds $\tau$}{
            continue\;
        }
        Add point-image pair to correspondence list\;
    }
}
\For{int i = 0; i $<$ num\_points; i++}{
    Initialize weight per category to 0\;
    \If{point has no correspondences}{
        continue\;
    }
    \For{int j = 0; j $<$ num\_correspondences; j++}{
        Compute the vote weight $w_{ij}$ as in Sec. 3.1.\;
        Add $w_{ij}$ to the category for this vote\;
    }
    Output the category with the highest weight\;
}
\For{int i = 0; i $<$ num\_points; i++}{
    \If{point has no correspondences}{
        Lookup nearest labeled neighbor in kdtree\;
        Output the category of the labeled neighbor\;
    }
}
\caption{Fusion Implementation Details}\label{alg:fusion}
\end{algorithm}

\subsection{Network Architecture Details}
\label{sec:appendix_architecture_details}

In this section, we provide details regarding the 2D3DNet network architecture used in our experiments, for reproducibility purposes.

\vspace*{2mm}\noindent\textbf{3D Architecture}
Our 3D network architecture is a 3D Sparse Conv Network~\cite{graham20183d}. Each encoder layer $(A,B)$ consists of a 3x3x3 spatial convolution with $A$ output features, followed by a 3x3x3 spatial convolution with $B$ output features, followed by a 2x2x2 3D maxpool operation. Each convolution is first normalized normalized by dividing by the activation of an occupancy feature convolution. Then batch norm is applied, followed by relu. The encoder layers are: (64, 64), (64, 96), (96, 128), (128, 160), (160, 192), (192, 224), (224, 256). The encoder layers are followed by a bottleneck layer. The bottleneck layer is (256, 256) and mirrors the encoder layers (the only difference is there is no maxpool after the convolution block). Finally there are decoder layers. Each decoder layer $(A,B)$ consists of a 2x2x2 unpooling, followed by a 3x3x3 spatial convolution with $A$ output features, followed by another 3x3x3 spatial convolution with $B$ output features. As before, each convolution is followed by relu and then by dividing by the activation of an occupancy feature convolution for normalization. The decoder layers are: (256, 256), (224, 224), (192, 192), (160, 160), (128, 128), (96, 96), (64, 64). Finally we apply a sequence of three 3x3x3 spatial convolution layers. The layer sizes are (64, 64, \texttt{output\_channel\_count}). The first two convolutions are followed by normalization, batch norm, and relu, as usual, while the last convolution is followed only by normalization. Finally we apply a softmax function to get probabilities (we use a softmax cross entropy loss). The loss is applied to voxels. For inference, we project back to points. At training time, we pad to 120,000 voxels. We randomly sample 1M points from the scene for input to the network for memory reasons. Where not otherwise specified, the input features are a one-hot encoding of the predicted semantic class per point (no other features).

\vspace*{2mm}\noindent\textbf{3D Optimizer and Data Augmentation Details}
We use a momentum optimizer with 0.9 momentum and an initial learning rate of 0.03. We use this learning rate for 200K steps. After 200K steps, we begin a cosine decay from 200K steps to 700K steps, at which point the learning rate fully decays to 0 and training is complete. The loss is softmax cross entropy classification loss plus a weight decay of 0.0001. We use 20cm voxels and a batch size of 8. For data augmentation, we apply the following at training time only: random crops of 3 square meters, x and y rotations of up to +-10 degrees, z rotations of up to +-180 degrees, and a random scale factor between 0.9 and 1.1.

\vspace*{2mm}\noindent\textbf{3D Details (nuScenes)}
For nuScenes, the 3D pipeline is identical to our dataset except for the following specific changes. First, the voxel size is reduced from 20cm to 5cm, training time voxel padding is increased from 120,000 to 240,000, and the batch size is reduced from 8 to 3. Second, no random crops, rotations, or scales are done (i.e., we skip the data augmentation). Finally, we train for only 450K steps rather than 700K; there is still a 200K warmup period with the same learning rate, but the cosine decay schedule is accelerated to reach 0 at 450K. Note that in addition, when doing multiview fusion on nuScenes we find and cull ego vehicle points when making the depth images, as otherwise the ego vehicle unnecessarily occludes a portion of the final image.

\vspace*{2mm}\noindent\textbf{2D Architecture}
Our 2D network architecture is DeepLabv3~\cite{chen2017rethinking} with Xception~\cite{chollet2017xception}. For our dataset, we use Xception65 and source images downsampled to a resolution of 1152x768. For nuImages, we use Xception71 and images of resolution 1600x900. We pretrain the network on ImageNet classification~\cite{ILSVRC15}. On nuImages we optimize with a batch size of 32, a backbone output stride of 16, a decoder output stride of 4, a base learning rate of 0.0045, 150K total steps, atrous rates of 6, 12, and 18, apply a hard pixel mining loss on the top 0.25\% of pixels, and set a last layer gradient multiplier of 10.

\subsection{Normal and Radii Estimation}
\label{sec:normal_estimation}
Here, we describe our approach to estimating surface normals and radii per lidar point. Many other methods of normal estimation would likely be fine, especially since their primary use in our pipeline is to create depth maps. We chose this particular approach because we found it generated high quality results on our data. Conceptually, we estimate a coordinate frame per point by doing a PCA decomposition on a neighborhood around the point, growing the neighborhood until the PCA decomposition is nondegenerate. Then the tangent, bitangent, and normal are the first, second, and third principal axes, respectively. Radii along the tangent and bitangent directions are set to a fixed fraction of the size of the neighborhood needed to get a stable decomposition. This defines a surface disk (surfel) parameterized by a normal, two radii, and a tangent.  Implementation details are below.

Consider one lidar point. First, we start by creating a neighborhood around the point with radius 25cm. Next, we use a kdtree to gather all points in the neighborhood. If there are too many points ($>32$), we randomly subsample until there are at most 32. We then do a PCA decomposition of the remaining points centered at the point in question. If there are not enough points to do a PCA decomposition ($<3$) or the PCA is degenerate, we double the neighborhood radius and try again, up to a maximum radius of 2m (if there are still not enough points for a nondegenerate PCA decomposition, we give up and simply set the normal to the world up vector, though this is exceedingly rare in practice). We determine degeneracy for the decomposition by asserting that the first two principal standard deviations are at least 10\% of the neighborhood size, clipped to the narrow range $[.25\mathrm{cm},1\mathrm{cm}]$.
Once the PCA is determined nondegenerate, it defines a coordinate frame around the point based on the orthogonal directions of maximum variance. The direction of maximum variation is chosen as the tangent, and the radius in the tangent direction is set to 0.25 * the neighborhood size. The second radius, the radius in the bitangent direction, is set to the aspect ratio * the first radius. The aspect ratio is determined as the ratio of the standard deviations of the first two principal axes (i.e., $\sqrt{\frac{\sigma_1}{\sigma_0}}$).

\subsection{Optimization Details}

In this section, we report details for the optimization algorithm described in Section~\ref{sec:sampling}.  The problem is to solve:
$$ 
\max_{S \subset \mathcal{I}} \quad  \norm{ \sum_{I \in \mathcal{S}} v_I}_{\frac{1}{2}}^{\frac{1}{2}}
\hspace{10mm}
\textrm{s.t.} \ |S| = N
$$ 

Before beginning the optimization, we compute terms $v_I$. For each image, we run our pretrained 2D conv network. We compute a histogram of class frequencies, and set $v_{I,c} := 1$ if at least 2\% of the predicted pixels for $I$ are class $c$, and 0 otherwise. Once $v_I$ is computed for all images, we run the optimization. There are  1.04 billion images in the candidate set, so efficiency is critical. In order to make it tractable, we separate the optimization into two phases. First, we filter out images that do not contain at least one fairly rare class. This removes the vast majority of images while being unlikely to substantially affect the final solution quality; it also makes it possible to fit all remaining $v_I$ in memory. The second step is a greedy algorithm. We start with an empty set, and repeatedly iterate through the remaining set of candidate images, choosing the image that most improves the current solution. At each step we add the chosen image to the current solution, and recompute the current overall energy. Then, we iterate over each image and compute the affected partial terms of each remaining image's nonzero categories with respect to the objective.  Once we have reach $|S| = N$, the optimization ends. This approach is not guaranteed to give an optimal result, but is necessary to accommodate the large number of image vectors.

\subsection{Evaluation Details}
\label{sec:evaluation_details}

In this section, we provide details about how
ground truth labels were acquired and used for evaluation in the twenty city dataset.

Our 2D image collection has labels for 44 classes.  There are no currently available 3D semantic labels. In order to evaluate our method in 3D, we annotate 20 scenes (approximately 50 million LiDAR points), directly adapting the class definitions from 2D. As described in Section~\ref{sec:dataset}, there are some differences in 2D versus 3D labeling standards. For example, in the 2D dataset, electrical wires were not labeled; we label wires separately and do not evaluate at those points.

Labeling the classes in 3D for evaluation poses some challenges. Some of the classes, such as vehicle on rails, are quite rare and therefore unlikely to appear repeatedly in a small number of scenes. Some classes, such as street light (which contains only the bulb, not the attached pole), are small. This makes them more difficult for a human to reliably find and segment accurately, and increases the effect of noisy boundaries on the metrics. Some classes, like animal or parking meter, are both small and rare, compounding these issues. Finally, some classes (e.g. sky, water) do not appear at all in the 3D point cloud. Therefore, in Section~\ref{sec:comparisons} we only report results for the 21 classes that appear in at least three different scenes with at least 1,000 points in each scene. This is 99.34\% of all scanned points.

In total, we drop 0.66\% of the 50 million points from the evaluation when reporting the mean IoU, most of which are large objects from classes that appear rarely (e.g., bridge). The classes that are kept versus those that are removed are provided in Table~\ref{tab:full_quant}.

It is important to note that while the labels that are thrown out are less reliable than those used for evaluation, not all methods perform equally well on the noisy classes. One important bias is that our 3D network is based on sparse voxel convolutions with a resolution of 20cm$^3$. As a result, our method is not well suited for objects that are small compared to this resolution. Fusion approaches are comparatively less affected by small objects. An extreme case is the fire hydrant class. This class accounts for only 0.004\% of 3D points, but our MVFusion approach still achieves an IoU of 60.4; however, all tested methods based on sparse voxel convolutions achieve an IoU of zero, a difference well in excess of the noise in the GT for this class. In order to handle these classes robustly, it would be necessary to adjust the 3D approaches to compensate for this deficiency.

{\small
\bibliographystyle{ieee_fullname}
\bibliography{egbib,city}
}

 \end{document}